\documentclass[lettersize,journal]{IEEEtran}
\usepackage{graphicx,color}
\usepackage{textcomp}
\usepackage{xcolor}
\usepackage{hyperref}
\hypersetup{hidelinks=true}
\usepackage{algorithm,algorithmic}

\usepackage{amsmath,amssymb,amsfonts}
\usepackage{array}
\usepackage{cite}

\usepackage{textcomp}
\usepackage{stfloats}
\usepackage{url}
\usepackage{verbatim}
\usepackage{graphicx,color}
\usepackage{calc}
\usepackage{xcolor}
\usepackage{tikz}
\usepackage{transparent}
\usepackage{subcaption}
\usepackage{placeins}
\usepackage{pgfplots}
\usetikzlibrary{patterns, arrows.meta, decorations.pathreplacing,positioning, fit, backgrounds, calc}
\pgfplotsset{compat=1.18}
\def\BibTeX{{\rm B\kern-.05em{\sc i\kern-.025em b}\kern-.08em
    T\kern-.1667em\lower.7ex\hbox{E}\kern-.125emX}}
\AtBeginDocument{\definecolor{tmlcncolor}{cmyk}{0.93,0.59,0.15,0.02}\definecolor{NavyBlue}{RGB}{0,86,125}}

\begin{document}


\newcommand{\leftTrim}{576}   
\newcommand{\rightTrim}{672}  
\newcommand{\topTrim}{346}    
\newcommand{\bottomTrim}{86}  

\title{A novel autonomous microplastics surveying robot for beach environments}
\author{Hassan Iqbal, Kobiny Rex, Joseph Shirley, Carlos Baiz, Christian Claudel
\thanks{This work is supported by Welch Foundation (F-1891), Matagorda Bay Mitigation Trust, and the UT Austin Office of the Vice President for Research, Scholarship and Creative Endeavors.}
\thanks{Corresponding author: H. Iqbal is with the Oden Institute for Computational Engineering \& Sciences. This research was conducted at the Department of Civil, Architecture and Civil Engineering, The University of Texas at Austin, USA (e-mail: hassan.iqbal@utexas.edu).}
\thanks{C. Claudel is with the Department of Civil, Architecture and Environmental Engineering, The University of Texas at Austin, USA (e-mail: christian.claudel@utexas.edu).}
\thanks{J. Shirley, K. Rex, C. Baiz are with Department of Chemistry, The University of Texas at Austin, USA (e-mail: cbaiz@cm.utexas.edu, \{joseph.shirley, kobyrex\}@utexas.edu).}}

\maketitle
\begin{abstract}
Microplastics, defined as plastic particles smaller than 5 millimeters, have become a pervasive environmental contaminant that accumulates on beaches due to wind patterns and tidal forcing. Detecting microplastics and mapping their concentration in the wild remains one of the primary challenges in addressing this environmental issue. This paper introduces a novel robotic platform that automatically detects and chemically analyzes microplastics on beach surfaces. This mobile manipulator system scans areas for microplastics using a camera mounted on the robotic arm's end effector. The system effectively segments candidate microplastic particles on sand surfaces even in the presence of organic matter such as leaves and clams. Once a candidate microplastic particle is detected, the system steers a near-infrared (NIR) spectroscopic sensor onto the particle using both NIR and visual feedback to chemically analyze it in real-time. Through experiments in lab and beach environments, the system is shown to achieve an excellent positional precision in manipulation control and high microplastic classification accuracy. 
\end{abstract}



\section{Introduction}
\label{sec:introduction}
Microplastics, generally classified as plastic particles less than 5 millimeters in size, have emerged as a significant environmental concern due to their physical and toxicological effects on a wide range of marine life. Beaches serve as major accumulation zones since they collect plastics originating from both land and water bodies. Additionally, beaches are dynamic systems, with constant movement of sand, shells, glass and plastic particles. Therefore, the detection, tracking, mapping and classification by chemical composition of microplastics across beaches is crucial to understand the phenomena and mechanisms associated with plastic deposition and spread, map pollution levels and inform public policy (plastic usage and disposal).


Current microplastic mapping processes involves two phases, and are largely non-automated. The first fieldwork phase includes three stages: location of high tide line and selection of transect, random selection of sampling points and sampling (collection of sand samples). The second phase, typically performed in the laboratory, includes sand drying, manual microplastic extraction, classification and quantification. Labor-intensive techniques such as sieving, density separation, and spectroscopic analysis of individual particles (e.g., Fourier transform infrared spectroscopy) are employed for this second stage. We refer the reader to the survey \cite{alvarez2020method} for a detailed account on current techniques for microplastics extraction and analysis.

The current process of microplastics sampling on beaches is labor-intensive and time-consuming. Samples are extracted using stainless steel tools either in situ or in the lab. The former is known to increase cross contamination risk while the latter involves transporting materials in zip-locked plastic bags to the lab where they are analyzed. The time-consuming nature of sampling makes it difficult to map spatio-temporal distribution (by chemical compositions) of microplastics on beach surfaces, as well as to identify their accumulation zones, potential sources, and concentration gradients.

The development of autonomous ground rovers for microplastic detection is hindered by several challenges. First, in situ chemical analysis of microplastics on beaches requires a high signal-to-noise ratio (SNR) in order to overcome environmental noise and identify subtle spectral differences between materials. In particular, the SNR must be sufficiently high to handle positioning errors in mobile manipulation (due to rough terrain, linearization assumptions in visual servoing, etc.) and to reliably differentiate partially covered in sand microplastics from nearby organic matter, shells, and sand particles. While hyperspectral cameras can capture the entire scene at once and mitigate positioning precision requirements, such cameras that cover broad enough spectra range with required SNR and resolution remain prohibitively expensive, bulky, and power-intensive for autonomous mobile robotics applications. On the other hand, off-the-shelf portable spectrometers can only analyze a single point at a time and would require significant human involvement in the process, in order to precisely position the plastic particle at the focal point of the spectrometer.

To address these gaps, we propose a field autonomous ground rover for real-time detection, quantification and classification by chemical composition of microplastics on beaches in situ. The key idea is to develop a visual servoing and near-infrared (NIR) based control for precision mobile manipulation. A NIR lamp is used to illuminate candidate microplastic particles, and the scattered light is analyzed using spectrometry to classify microplastics.

\begin{figure*}[!htbp]
    \centering
    \begin{tikzpicture}
        \node[inner sep=0pt] (base) at (0, 0) {
            \begin{minipage}{\textwidth}
                \centering

                \begin{minipage}{0.328\textwidth}
                    \centering
                    \includegraphics[width=\textwidth,trim={680 250 680 250},clip]{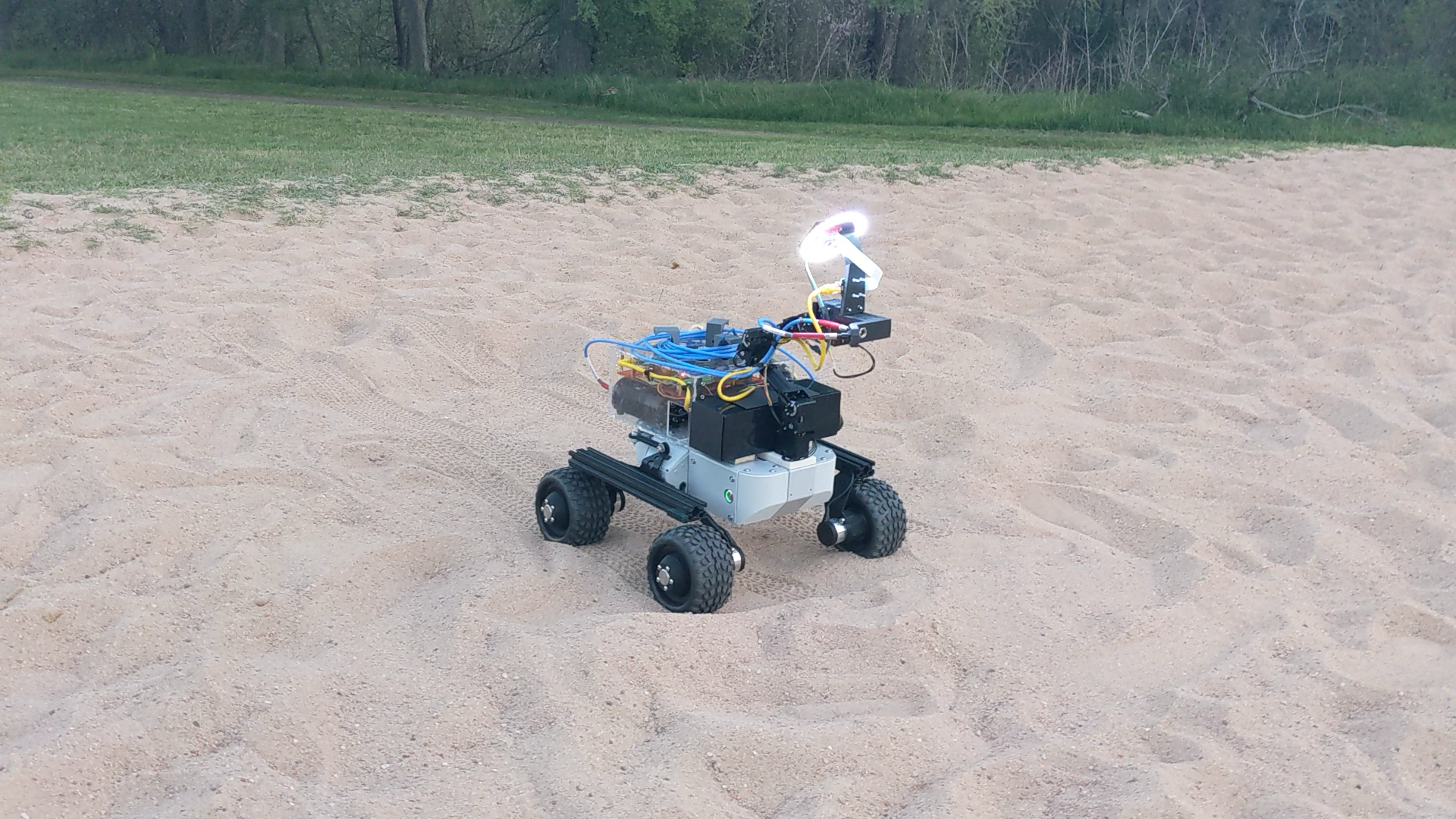}
                \end{minipage}
                \hfill
                \begin{minipage}{0.328\textwidth}
                    \centering
                    \includegraphics[width=\textwidth,trim={500 140 860 360},clip]{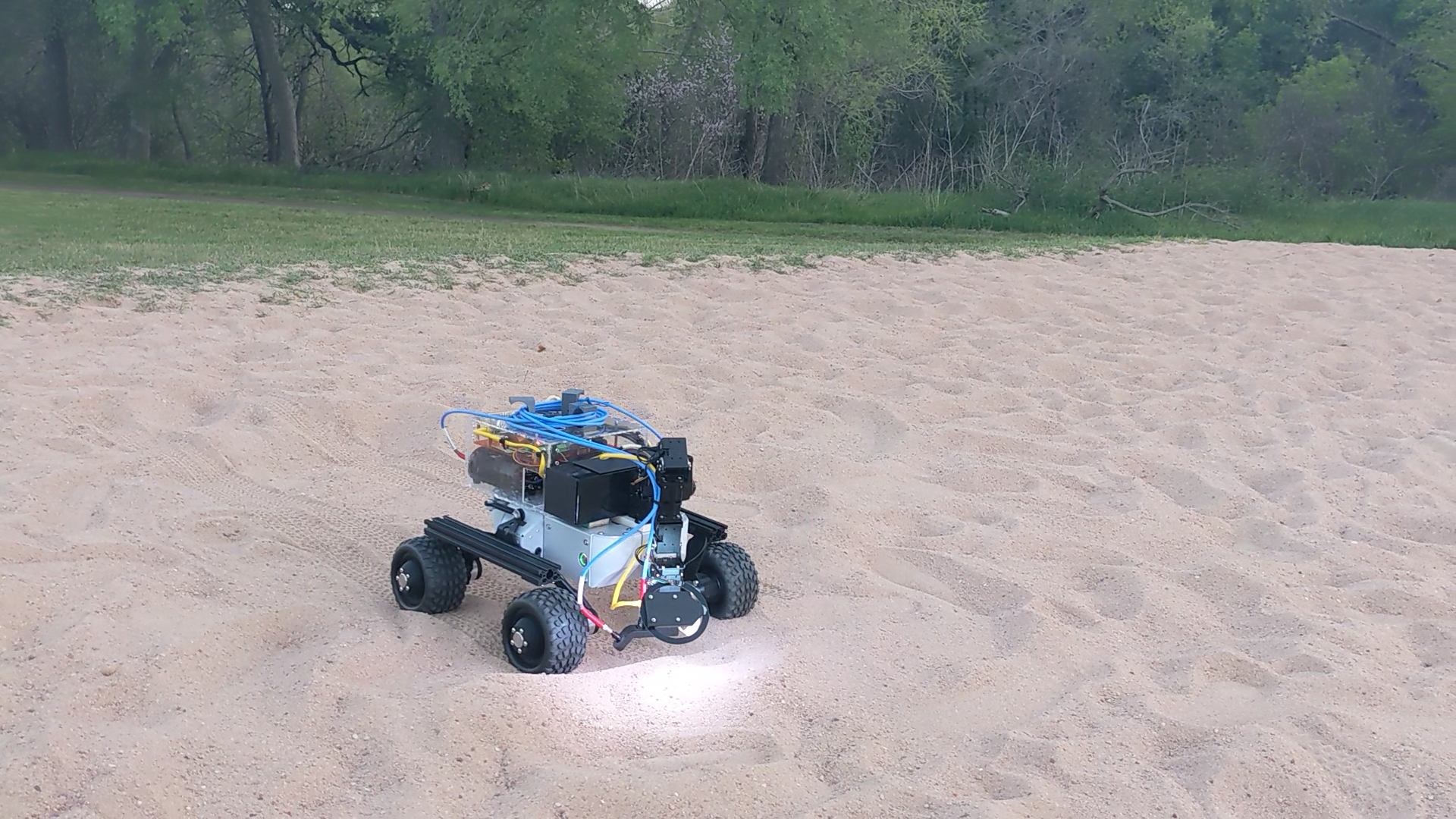}
                \end{minipage}
                \hfill
                \begin{minipage}{0.328\textwidth}
                    \centering
                    \includegraphics[width=\textwidth,trim={520 320 840 180},clip]{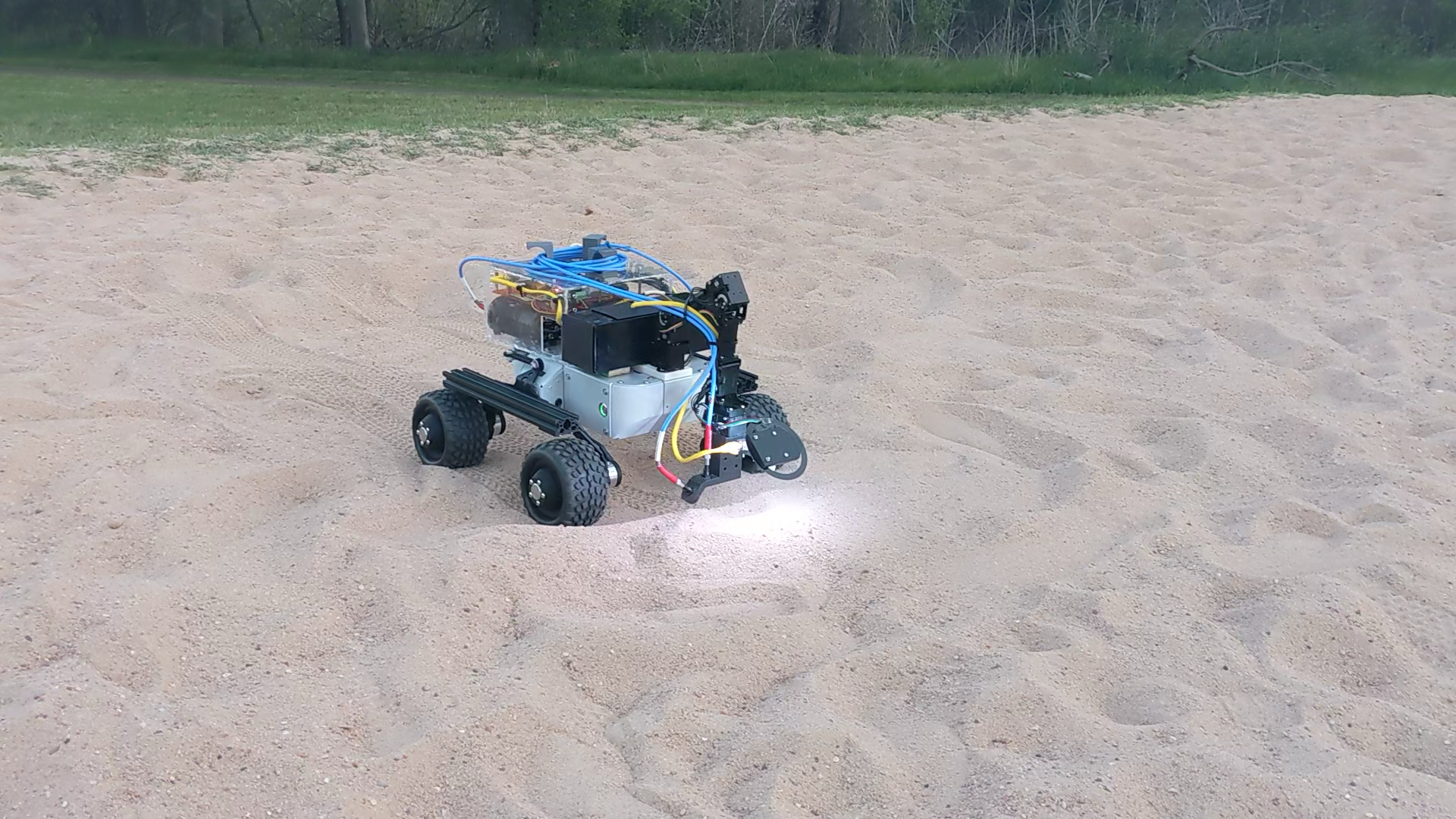}
                \end{minipage}
            \end{minipage}
        };
        \node at ([yshift=-9.25cm]base.north) {

        \includegraphics[width=\textwidth,trim={0 336 307 547},clip]{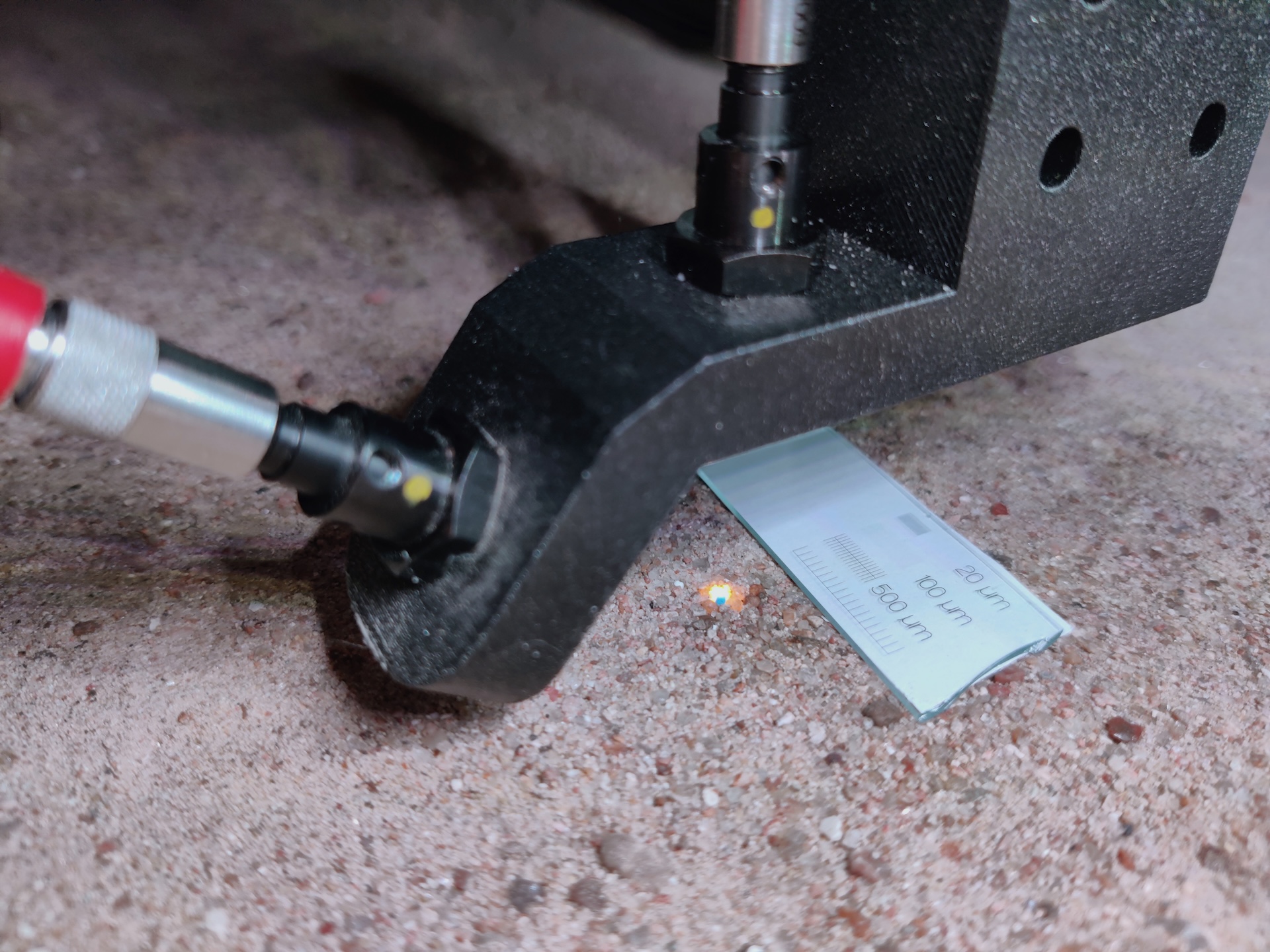}
        };

        \node[opacity=0.75,anchor=north west] at ([xshift=0.01cm,yshift=-7cm]base.north west) {
            \includegraphics[width=0.32\textwidth]{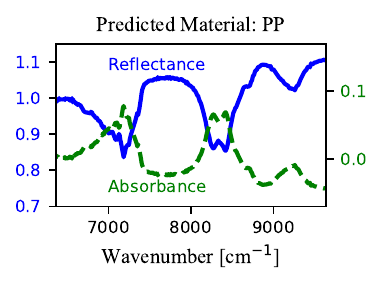}
        };

    \end{tikzpicture}
    \caption{Top left: Undone navigation with the robotic arm's pose set to maintain center of gravity in a position that prevents the rover from tipping over. Top center: Space around the mobile base is being scanned with camera mounted on robotic arm's end effector. Top right: A candidate microplastic particle is detected and robot is ready to track the particle for chemical analysis. Bottom: Close-up view of end-effector. Particle is tracked and NIR illumination is focused on top of a blue microplastic. Corresponding absorbance and reflectance spectra from raw (noisy) measurements are plotted against wavenumbers. The machine learned classifier correctly identifies the particle as polypropylene (PP) based on its distinctive spectral signature. A ruler is added for visualization purposes.}
    \label{fig:demo-sequence}
\end{figure*}

Distinguishing microplastics from similarly-sized sand grains in unstructured, dynamic environments presents a significant detection challenge in its own right. The extreme class imbalance between microplastics (rare positives) and sand particles (abundant negatives) complicates segmentation tasks using conventional machine learning models. Additionally, illuminating objects with NIR light (with a reddish hue) alters their apparent colors, further compounding visual detection and the meta problem of visual servoing for precise robot control. Hence, the focus of this work is to develop a novel visual servoing approach that leverages NIR properties to realize millimeter-scale positional errors in unstructured environments. Our key contributions include:

\begin{enumerate}
    \item Design of a novel mobile manipulator with end effector that integrates high resolution camera and near infrared spectroscope, that allows for precise targeting and spectral analysis of millimeter-scale microplastics particles despite surface irregularities and platform instabilities inherent to beach environments;
    \item A specialized image-based visual servoing method that precisely positions the spectroscopic sensors for a reliable \textbf{chemical analysis} of microplastics under \textbf{occlusion};
    \item A real-time classification algorithm capable of identifying microplastics types by their chemical compositions from NIR spectral data, while remaining robust to systemic and random uncertainties;
    \item Experimental validation demonstrating successful detection and classification of microplastics in beach environments.
\end{enumerate}

Although outside the scope of this work, the proposed solution is designed to help improve the understanding of beach dynamics, such as the identification of loci associated with higher concentrations of microplastics within beaches and the relationship between microplastics abundance and sediment grain size. A richer and faster sampling enabled via proposed platform shall provide concentration gradients that allow for timely identification of microplastics sources and fate in beaches \cite{graca2017sources}. Similarly, temporal variations in concentration provide important information on on the impact of tides and on the effects of high-wind events. We defer these to future work. In fig. \ref{fig:demo-sequence}, the developed platform's operation is shown detecting and classifying a blue microplastics particle on a beach surface.

\section{Related Work}
\label{sec:related_work}


\subsection{Environmental exploration and autonomous science} Autonomous mobile robots, used as a single or multi agent system, have become vital for large-scale environmental monitoring. Recent advances have enabled precise data collection across land, sea, and air \cite{dunbabin2012robots}. There has been renewed interest in automating science labs so that tedious experiments can be performed with higher accuracy, efficiency and repeatability. \cite{angelopoulos2024transforming}. This work is at the intersection of autonomous science and environmental exploration.

\subsection{Desk-top precise placement} Prior research has demonstrated autonomous precise placement in various contexts. Angelopoulos et al. designed a robot to autonomously place injection needles for gas chromatography using multiple cameras mounted on the workbench \cite{angelopoulos2023high}. Several studies have developed robotic arm controllers for precise ultrasound probe placement by leveraging real-time ultrasound image feedback and visual servoing \cite{sauvee2008ultrasound,ma2024guiding,mebarki20102,nadeau2013intensity,nadeau2016moments,pane2022ultrasound}. Such studies often utilize eye-to-hand visual servoing. This relies on mounting cameras/ultrasound probes such that their field-of-view (FOV) always covers the entire workspace. While our work also focuses on precise probe placement, it addresses unique challenges: unstructured environments with uneven terrain and a robotic arm mounted on a mobile base (mobile manipulator). Furthermore, our application requires long-range environmental exploration and a relatively small-sized robot for maneuverability reasons, making it impractical to mount external cameras with a field of view covering the entire workspace. This limitation rules out eye-to-hand visual servoing approaches. We develop a eye-in-hand technique where the camera is mounted directly on the robot's end-effector. Additionally, we utilize near-Infrared (NIR) spectrometry, which provides point measurements rather than image-based data used in ultrasound applications. 

\subsection{Planetary robotics} Researchers have developed algorithms for rover-mounted manipulator arms that accurately place an instrument on a target designated from a few meters away. Planetary robotics typically focuses on contacting rock core and soil samples for spectroscopy \cite{baumgartner2006mobile,bowkett2024challenges}. Backes et al. developed a stereo-vision based rover and instrument placement algorithm \cite{backes2005automated}. Fleder et al. showed the autonomous rover traverses rocky terrains and achieves terminal precision of 2-3 cm from target designated with point-and-click mouse commands \cite{fleder2011autonomous}. Compared to planetary robotics, our problem requires working at considerably low-power spectrometry and low-weight for safe operation in public beaches. Unlike planetary robotics applications, we also require distinguishing microplastics from organic matter and sand particles of varying size and moisture levels. For safety reasons, we also exclude sensing systems that could pose hazards to humans or wildlife, such as high power UV lasers used in remote Raman spectroscopy applications. An initial version of the proposed robot used Raman spectroscopy, but this version was abandoned given UV eye safety constraints that considerably slowed down the classification process. 

\subsection{Manipulators with spectrometers} In \cite{hanson2023slurp}, materials such as oil, water and sugar materials are classified by grasping their sample container using a gripper on the arm and analyzing them with spectrometers. The work \cite{hanson2024prospect} proposed planner for a table-top mounted robotic arm to autonomously place fiber optic cable on material surfaces in order to classify them as sandstone slabs, Gypsium boulder etc. However, these studies have been performed in structured environments and objects of interest are significantly larger in size as compared to microplastics. 

\begin{figure}[t]
    \centering
        \includegraphics[width=0.5\textwidth, trim = {0 10pt 0 10pt},clip]{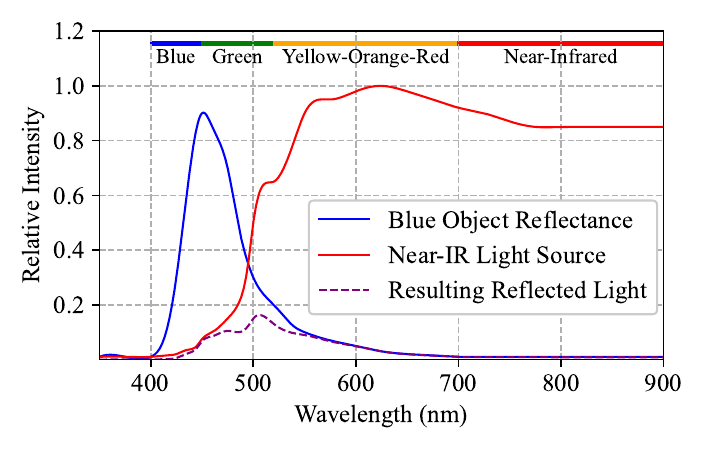}
    
        \caption{Spectral density plot showing why blue objects appear black under near-infrared illumination.}
    \label{fig:spectral_density}
\end{figure}

\subsection{Microplastics segmentation} There has been limited but some work on segmenting microplastics fragment from their backgrounds by training deep learning models on microscopic images \cite{huang2023proceeding,park2022mp,xu2024efficient,phan2023exploiting}. Microscopic images used in these studies do not have the stark class imbalance as in raw images captured from ground rover operating on the beach. Furthermore, images collected during rover operation on the beach may have complex backgrounds with clams, leaves and sand particles, lighting may vary and the microplastic particle may not be perfectly focused. Our work addresses these issues. Namely, for successful segmentation in the field, we develop three separate but interacting routines: intensity-based thresholding, camera focusing, and controlled illumination with programmable LEDs. Furthermore, we develop a NIR signal-based classifier capable of identifying various microplastic types \cite{shirley2025microplastics} (including Polypropylene (PP), Polystyrene (PS), Polyvinyl Chloride (PVC), Polyethylene Terephthalate (PET), Nylon and others) even when particles are wet or intermixed with organic plant material. The meta-routine, which integrates segmentation and inference from the learned classifier within manipulator control loop, enables us to analyze microplastics with an excellent success rate.
\subsection{Occlusion in visual servoing} The goal is to control the rover-mounted manipulator using real-time visual images and NIR measurements. This has intricate connection with visual servo control for robotic manipulators. Similar to \cite{dong2016incremental,kelly2000stable}, we compute incremental joint angles to use as control inputs. Crucially, this avoids multiple solutions in the existing inverse kinematics, as well as performance issues caused by errors caused by miscalibration of vision system or poor estimation of kinematic model parameters of the mobile manipulator \cite{wijesoma1993eye}. We also face the issue of occlusion when the NIR illumination is close to microplastic particle. For example, fig. \ref{fig:spectral_density} shows this effect using spectral density analysis i.e. blue microplastics would exhibit a dramatic hue transformation, appearing almost completely black under incandescent bulb lighting used for NIR illumination. Occlusion in visual servoing is handled in agricultural environments with multiple cameras \cite{yoshida2022automated,bajracharya2006vision}, planning the next best view \cite{sun2024efficient,border2024surface,radmard2018resolving}, out-painting the end effector \cite{gupta2025training} and predicting image features that are out of the camera FOV \cite{xin2021visual}. To handle our occlusion problem, during the terminal state i.e. when current pose is close to designated target, we servo manipulator using both NIR point measurements and visual feedback. 


\section{Robotic Platform}

\subsection{System design and assumptions}
\subsubsection{General design considerations}
This robotic platform is built around a four wheeled, small scale robot. Mechanical, optical, computing and sensor choices are made to allow for stability on beach surfaces and low energy consumption per unit distance. The robot is designed to operate at night to ensure high SNR and consistent lighting conditions. We program the sampling step in function of observed concentration gradients, with smaller steps in areas with higher variability. 

\subsubsection{Sensing apparatus}
The fastest method to localize potential microplastic particle candidates is by using visual imagery captured through high-resolution cameras. However, candidate particles need to be further analyzed to know if these particles are indeed plastics: some microplastics such as tire particles or white plastic particles are very similar to naturally occurring dirt or shell debris. We also need to classify them into different plastic categories: this classification process help understanding the origin of plastic, and the type of object that broke down to create the observed microplastic. To perform this, we use the ``NIR Quest'' spectrometer that allows an excellent signal-to-noise ratio of 15,000:1. Since the spectrometer takes point measurements, a very high positioning precision (mm scale) is required for the measurement apparatus with respect to the candidate particle. 
\begin{figure}[ht]
    \centering
    \includegraphics[width=\columnwidth]{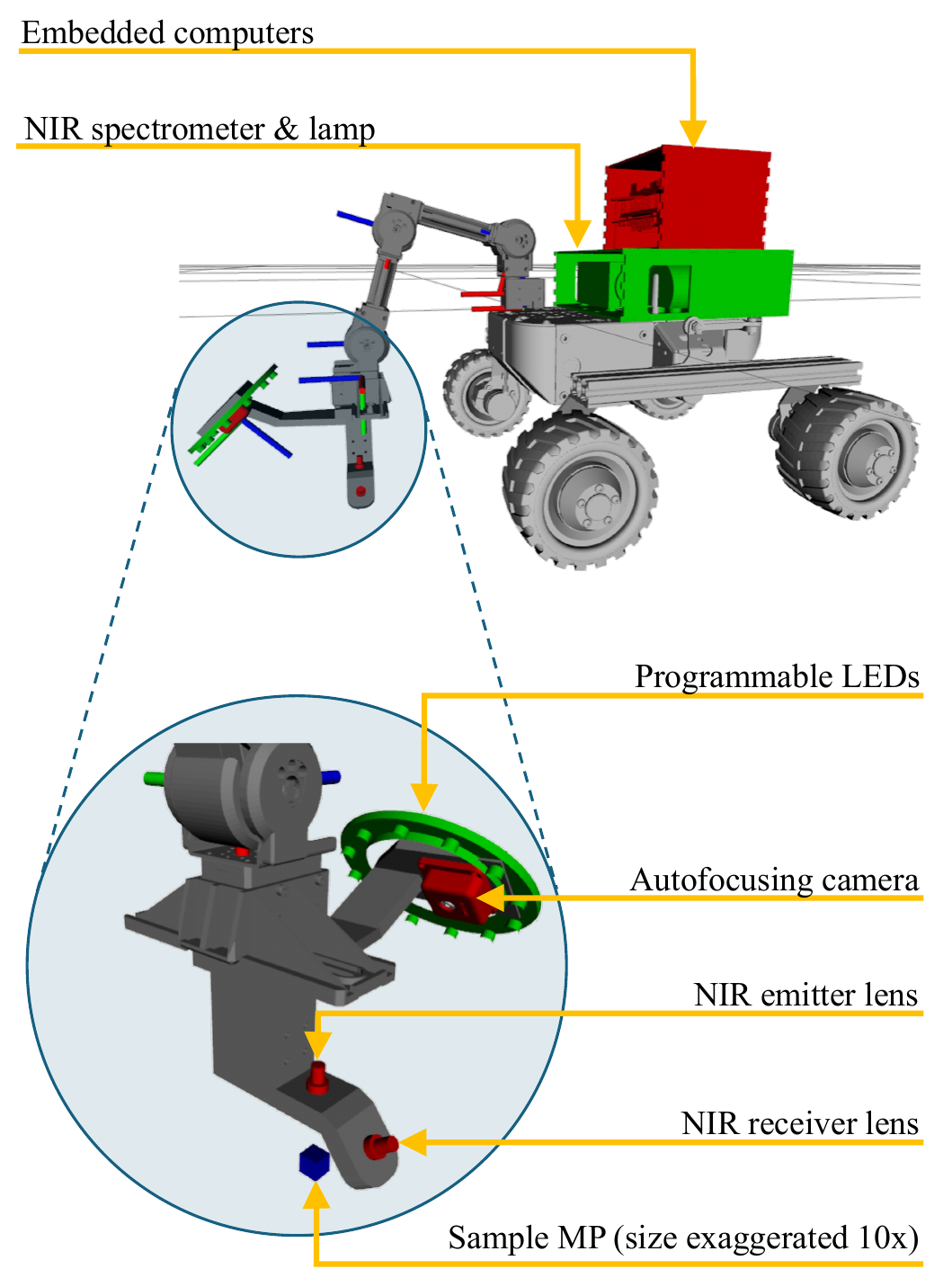}
    \caption{End effector design and a candidate microplastic particle ``MP sample'' shown in blue. Size exaggerated for visualization purposes.}
    \label{fig:end_effector_design}
\end{figure}

In practice, this precision can be achieved with an appropriate robotic arm. We choose the OpenManipulator-X arm with four degrees of freedom due to its low weight and sub-mm repeatability. We replace the arm's default gripper with a custom-designed optical-NIR instrument that performs both detection and classification functions. The design of end-effector is shown in Fig. \ref{fig:end_effector_design}. 
\begin{figure*}[ht!]
    \centering
    \includegraphics[width=2\columnwidth]{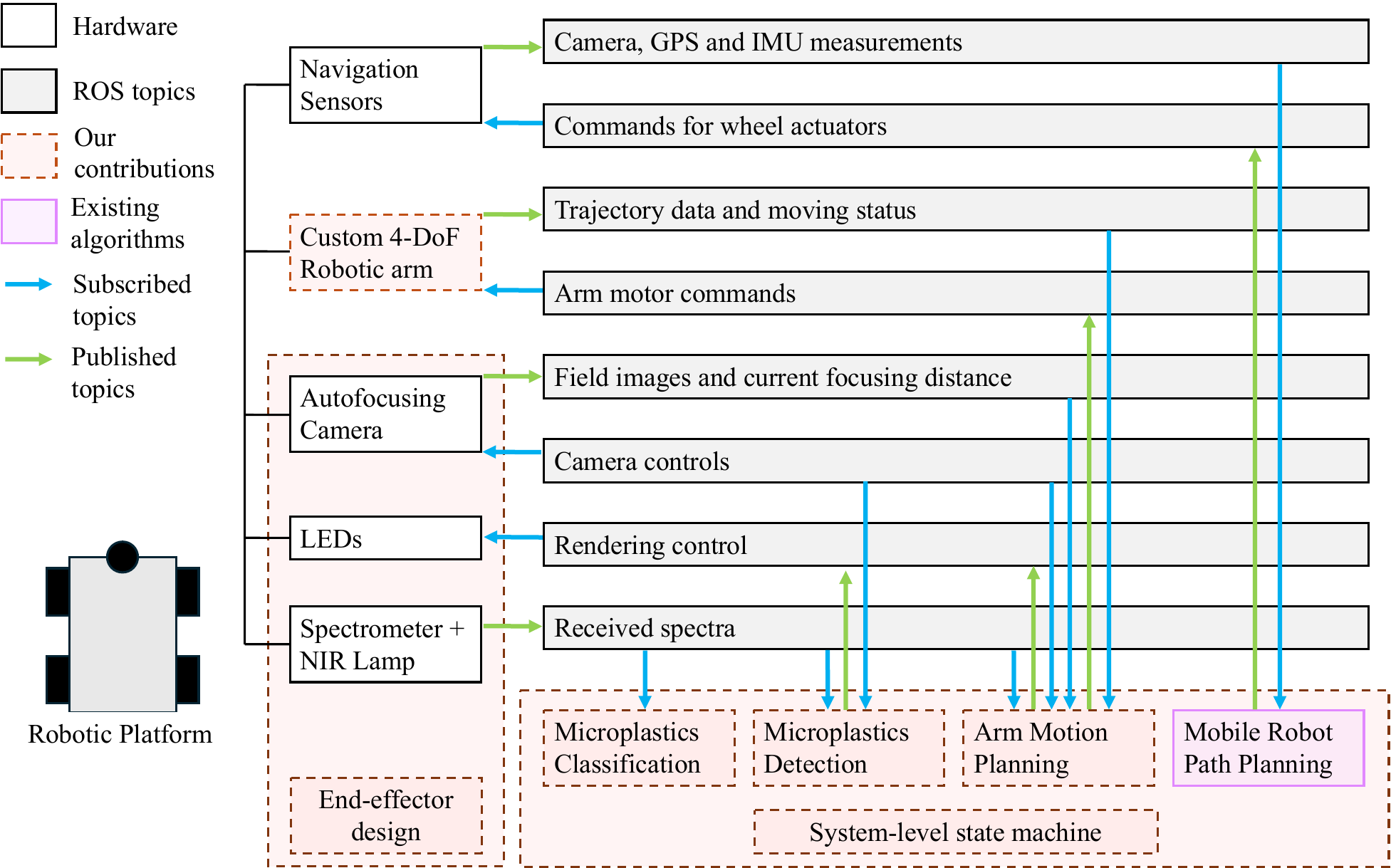}
    \caption{ROS topics and nodes in the robotic platform.}
    \label{fig:rostopics}
\end{figure*}
An incandescent light source (Ocean Optics HL-2000-LL-FHSA) is used to provide the required illumination for the near-infrared spectrometer, while a programmable LED ring is used to provide visible illumination for the main camera. As both the spectrometer and lamp are too bulky and fragile to be mounted directly onto the arm's end-effector, we design the end effector to instead mount two collimating lenses with focusing distance of 10 mm and supported wavelength range of 350 to 2000 nm. The emitter and receiver lenses are coupled to optical fibers which in turn are attached to lamp and spectrometer respectively. End effector is designed such that the relative poses of collimating lenses, camera and LED ring allow for clear view of beach surface, minimum impact of shadows, smooth and fine arm motions for precise point measurements.

\subsubsection{Mechanical considerations}

The robot is a four-wheeled robot of brand ``Leo Rover'' with mass $m=6.5$ kg, and dimensions $425\times448\times305$ mm. Two 45 Watt electric motors power the left and right wheels independently in order to both move and skid steer. The structure of the robot allows for a maximum total payload of 5 kg. Design specifications allow the robot to climb slopes up to 10 degrees before losing traction or tipping over. We position the battery, computers, spectroscope and actuators to allow for a 10 degree maximum longitudinal and cross slope, which is sufficient to cover most beach terrain. Since the robot is moving on soft sand, odometry is performed using image correlation/optical flow rather than direct wheel encoder measurements. 

\subsubsection{Power considerations}
The rover is powered by a 11.1 V 5.8 Ah battery while the robotic arm, spectrometer, lamp and embedded computers are powered by a separate rechargeable 12 V, 30 Ah Lithium nickel manganese cobalt (NMC) battery, weighing 4.6 lbs and capable of delivering a peak discharge current of up to 60 A (2C). We choose NMC Li-Ion as it allows for a higher energy density compared to more reliable Lithium Iron Phosphate (LFP) batteries. The power drawn by each subsystem is mentioned in the table \ref{tab:power_requirements}. Tests have shown that the battery lasts over three hours during regular operation. Combined with an average speed of 0.016 m/s, this allows the robotic platform to map microplastics over a 70 m$^2$  total area per charge. 

\begin{table}[h]
\centering
\begin{tabular}{lcc}
\hline
\textbf{Component} & \textbf{Voltage (V)} & \textbf{Current (A)} \\
\hline
Embedded computers & 12.0 & 1.25 \\
Spectrometer       & 5.0  & 3.00 \\
Lamp               & 12.0 & 1.20 \\
Robotic arm        & 12.0 & 5.00 \\
Rover drive system & 11.1 & 8.00 \\
\hline
\end{tabular}
\caption{Component-wise peak power requirements}
\label{tab:power_requirements}
\end{table}

\begin{figure*}[]
    \centering
    \includegraphics[width=2\columnwidth]{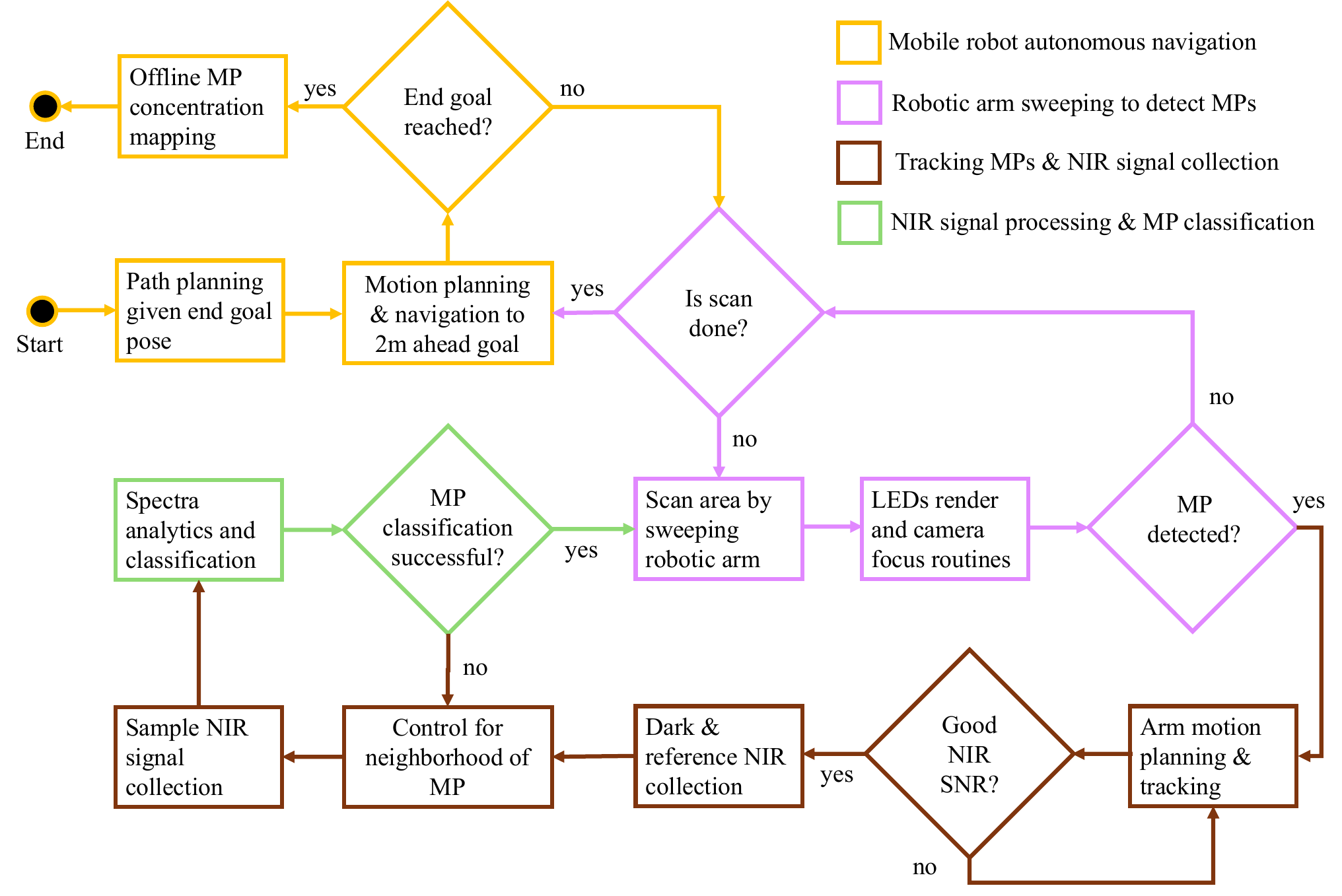}
    \caption{State machine based control.}
    \label{fig:state_machine}
\end{figure*}

\subsubsection{Embedded electronics and computational framework}

The robot is built around a main Nvidia Jetson Orin Nano computer that performs video processing, robot arm control, data analysis and robot steering. A dedicated microcontroller is used to control the LED light source for illuminating the scene. UART is utilized for communication between Jetson, microcontroller and drive controller. Sensors used to navigate the robot include wheel encoders, 9 DoF IMU, GPS, NIR spectrometer and a high resolution camera. 

We use and further develop open-source hardware and software to maintain reproducibility, modular design and future scalability. Namely, Robot Operating System (ROS) tools including topics, nodes, messages and services are utilized to develop and integrate compute nodes for each dedicated task. Nodes that subscribe and publish to various topics, as well as hardware and drivers developed in this work are shown in Fig. \ref{fig:rostopics}. Microplastics detection loop is run at 60 Hz while the higher level robotic arm planner is run at 10 Hz. Robotic arm low-level controllers are run at 100 Hz. NIR data acquisition to classify the type of microplastics takes 10 ms.





\subsection{Scanning and probe alignment}

\subsubsection{General sampling process}

The robotic platform is controlled via a high level state machine as shown in Fig. \ref{fig:state_machine}. The four main tasks include planning and navigation of the mobile base (rover); scanning the area around the base with robotic arm mounted camera (eye-in-hand configuration) and detecting potential microplastic candidates on sand surfaces; visual and NIR-based robotic arm control to steer NIR sensors onto candidate microplastic particles while ensuring an optimal SNR; NIR spectra acquisition for both reference (sand) and sample (candidate microplastic particle), NIR signal processing to get reflectance and absorbance signatures, microplastic particle type classification (or classifying it as a non-plastic object). Feedback position signals are derived from the high resolution camera and the spectrometer signal. We use both contrast and phase based feature detection to autonomously change the focusing distance of camera lens based on whether the current status is scanning the region around base or tracking a detected candidate microplastic particle when close to the ground ($\sim$10 cm). These main tasks are run in separate ROS nodes and their statuses are communicated to other nodes via ROS messages. 


\subsubsection{Arm control}

In this section, we explain the visual and NIR feedback loop to accurately control the pose of robotic arm's end effector relative to candidate microplastic sample. We defer the discussion on visual detection and tracking to section \ref{sec:visual_tracking}. 

We design a hierarchical control structure, with the vision system providing set points as inputs to the robotic arm's joint level controllers. In particular, we devise a dynamic image-based look-and move structure, according to taxonomy in \cite{hutchinson1996tutorial}. This approach makes our system portable and easily extensible to other robotic arms that allow interface for incremental position or joint velocities. Secondly, this allows for stable control as the robotic arm internal feedback loop generally runs at a higher rate (100 Hz in our case) than visual sampling rate (60 Hz in our case). As noted in \cite{hutchinson1996tutorial}, such a structure also decouples the visual controller from any kinematic singularities that are best handled with specialized resolved-rate robotic arm controllers. We compute control values based on image features directly (image-based servoing) to minimize computational delays and eliminate errors due to sensor modeling and camera calibration. This makes the underlying control a nonlinear and highly coupled problem. 

\begin{figure*}[htbp]
\includegraphics[width=2\columnwidth]{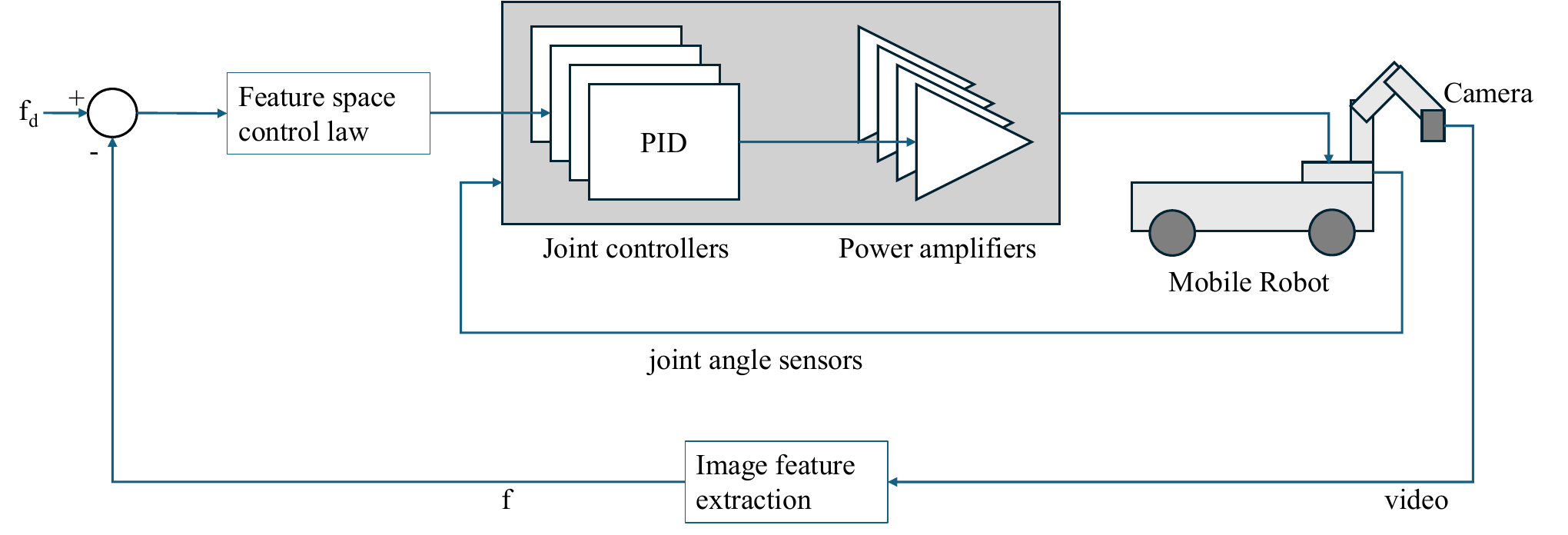}
\caption{Image-based visual servoing that observe both the target object and the robot end-effector as endpoint closed-loop (ECL) system.}
\label{fig:visual_servoing}
\end{figure*}

Image error function is defined as distance between candidate microplastics particle $f_d$ and the combined focusing point $f$ of emitter and receiver lenses on end effector. This error function, defined in image feature parameter space, is related to changes in joint coordinates $(\gamma_0,\gamma_1,\gamma_2,\gamma_3)$ and provided as control input to robotic arm joint controllers. A schematic of this process is shown in Fig \ref{fig:visual_servoing}. 

Through a camera calibration process, we determine intrinsic camera parameters such as focal length, pixel pitch and the principal point. Relative pose between camera and end effector is chosen during design of the custom end effector. Both extrinsic and intrinsic camera parameters are used to relate changes in image feature space to changes in joint coordinates. However, as mentioned previously, our image-based visual servoing mechanism allows a positional accuracy that is robust to miscalibration of vision and kinematic parameters. This is crucial for real-life experiments where systemic and random uncertainties may be present due to uncertain beach surface profile, lighting and other environmental conditions, unmodeled flex in arm due to wind or high payload, impact of rover suspension system on relative pose of robotic arm's base to rover's base, etc. 

To simplify the kinematics of the problem, we constrain the third joint vertical axis to remain in the same orientation as the vertical axis of the robot frame. We thus have $\gamma_1+\gamma_2+\gamma_3=\pi/2$, and thus set $\gamma_3=\pi/2-\gamma_1-\gamma_2$ in the remainder of the article. With this kinematic constraint, we only have to control angles $(\gamma_0,\gamma_1,\gamma_2)$. In practice, the arm mount slightly flexes under the weight of the arm, and thus the sum value is slightly adjusted from $\pi/2$ to ensure that the vertical axis of the third joint remains aligned with the robot vertical axis despite this flex for a range of nominal poses. In cylindrical coordinates $(r,\theta,h)$ associated with the rover frame, centered in the first joint of the arm, we have that:
\begin{equation}
\begin{array}{l} 
\theta = \gamma_0\\
r = r_1\cdot sin(\delta)+r_2\cdot cos(\beta)\\
h= r_1 \cdot cos(\delta)+r_2\cdot sin(\beta)\\
\end{array}.
\end{equation}

\noindent where $r_1$ is the radius of the first arm of the robot (linking joint 1 to joint 2), $r_2$ is the radius of the second arm of the robot (linking joint 2 to joint 3), $\alpha_j$ is a parameter associated with the robot arm design, $\delta = \gamma_1+\alpha_j$, and $\beta=-\gamma_1-\gamma_2$. We can write the coordinates of the last joint in Cartesian coordinates $(x,y,z)$ by defining the x-axis as the forward axis as:

\begin{equation}
\begin{array}{l} 
x = (r_1\cdot sin(\delta)+r_2\cdot cos(\beta))\cdot cos(\gamma_0)\\
y=(r_1\cdot sin(\delta)+r_2\cdot cos(\beta))\cdot sin(\gamma_0)\\
z= r_1 \cdot cos(\delta)+r_2\cdot sin(\beta)\\
\end{array}.
\end{equation}

Position feedback derived from the camera (for x-y coordinate control) and the spectrometer signal (for z coordinate control) is used to establish feedback control on the linearized model. With the above kinematics, we can compute the approximate position changes $(\delta x,\delta y,\delta z)$ associated with changes $(\delta \gamma_0, \delta \gamma_1. \delta \gamma_2)$ of control angles $(\gamma_0,\gamma_1,\gamma_2)$ as:

\begin{equation}
\left[\begin{array}{l} 
\delta x \\
\delta y \\ 
\delta z 
\end{array} \right] = J \cdot \left[ \begin{array}{l} \delta \gamma_0 \\
\delta \gamma_1 \\
\delta \gamma_2
\end{array} \right] 
\end{equation}

\noindent where J is the image Jacobian matrix associated with above-defined Cartesian coordinates, defined as


\begin{equation}
\small
J = \begin{bmatrix}
-(\phi_1 + \phi_2 c_\beta)s_{\gamma_0} & (\phi_3 + \phi_2 s_\beta)c_{\gamma_0} & \phi_2 s_\beta c_{\gamma_0} \\
(\phi_1 + \phi_2 c_\beta)c_{\gamma_0} & (\phi_3 + \phi_2 s_\beta)s_{\gamma_0} & \phi_2 s_\beta s_{\gamma_0} \\
0 & -\phi_1 - \phi_2 s_\beta & -\phi_2 s_\beta
\end{bmatrix}
\end{equation}

\noindent where $\phi_1 = r_1 s_\delta$, $\phi_2 = r_2$, $\phi_3 = r_1 c_\delta$, and $s_x = \sin x$, $c_x = \cos x$ for any angle $x$. Using the above linear approximation, we can compute at each step the theoretical changes in actuator angles $(\delta \gamma_0, \delta \gamma_1, \delta \gamma_2)$ as:

\begin{equation}
\left[ \begin{array}{l} \delta \gamma_0 \\
\delta \gamma_1 \\
\delta \gamma_2
\end{array} \right] 
 = \lambda  J^{-1} \cdot \left[\begin{array}{l} 
\delta x \\
\delta y \\ 
\delta z 
\end{array} \right] 
\label{eq:joint_angles}
\end{equation}

\noindent where $0<\lambda\le 1$ is a parameter that defines how large each displacement is, as a fraction of the total required displacement. For this particular setup, we choose a value of $\lambda=0.2$ which balances errors due to nonlinearities and errors due to uncertainty in pixel to distance relationships. Since the arm carries a relatively high payload, there is a relatively long (tens of seconds) settling time after each step. The total arm displacement speed is set relatively low as we require the arm to settle before another image is accepted for control.

Assuming camera is modeled by perspective projection, a point, $\textbf{P}= [ x_c ,y_c, z_c]^T$: whose coordinates are expressed with respect to the camera coordinate frame will project onto the image plane with coordinates $\textbf{p} = [u,v]^T$ , given by

\begin{equation}
    \begin{bmatrix} u \\ v \end{bmatrix} = \frac{\omega}{z_c} \begin{bmatrix}x_c \\ y_c\end{bmatrix}
    \label{eq:projection_eqs}
\end{equation}

\noindent where $\omega$ is a parameter associated with the camera. Since $z_c$ is approximately constant when we set the reference pose (it slightly varies in practice since the terrain is uneven), we can empirically determine $\omega/z_c$ which corresponds to a scaling parameter (in distance/pixel), and thus directly use image coordinates $[u,v]$ to estimate corresponding values of required displacements. We use the projection equations \eqref{eq:projection_eqs} and coordinate transformation from camera to the world frame in order to estimate $(\delta x, \delta y, \delta z)$. Then the required joint angle commands can be computed using eq. \eqref{eq:joint_angles}, and are sent to the arm controller. 

Once the effector has reached target positions for the (x,y) coordinates, we still have to adjust height z, since receiver and emitter lences are tuned to a precise focusing distance. Thus, we adjust the height using the signal generated by the spectrometer, starting with an initial arm altitude that is greater than the maximum required height (given expected potential perturbations in the terrain profile), and progressively move the arm down with small steps until the spectrometer signal exceeds a SNR threshold. To speed up the process, we also extract the diameter of the NIR illumination source spot using image segmentation. When the end effector is far from ground, this diameter is used to rapidly reduce height when the illumination spot is clearly out of focus (i.e. too large). 
\begin{figure}
    \centering
    \includegraphics[width=\linewidth]{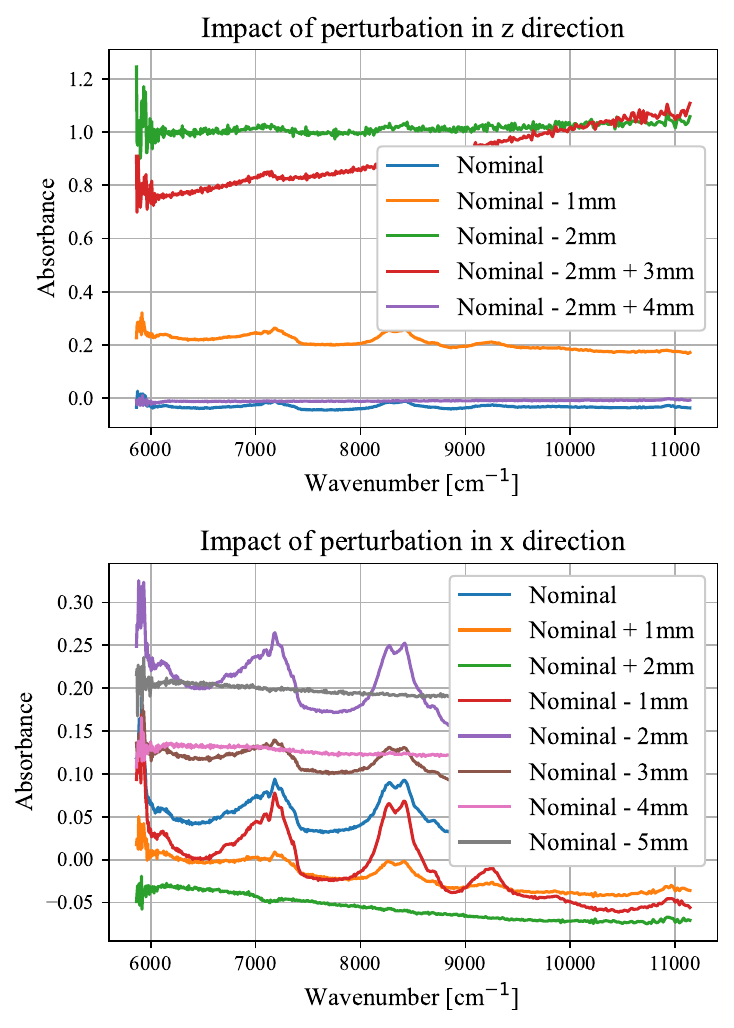}
    \caption{Absorbance profile for slightly perturbed NIR sensor placements to illustrate sensitivity and need for precision.}
    \label{fig:quantitative_spec}
\end{figure}

\begin{figure}
    \centering
    \includegraphics[width=\linewidth]{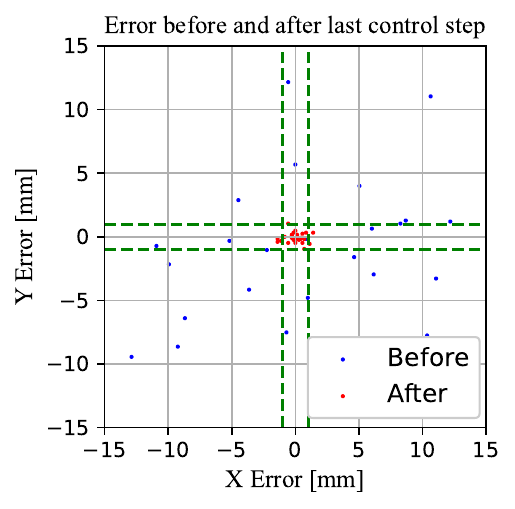}
    \caption{Random trials to evaluate errors before and after the terminal control step.}
    \label{fig:terminal_step}
\end{figure}

\subsubsection{Terminal control for occlusion}

\begin{figure*}[htbp]
    \centering
    \begin{minipage}{0.49\textwidth}
        \centering
        \includegraphics[width=\textwidth,trim=0 0 0 120,clip]{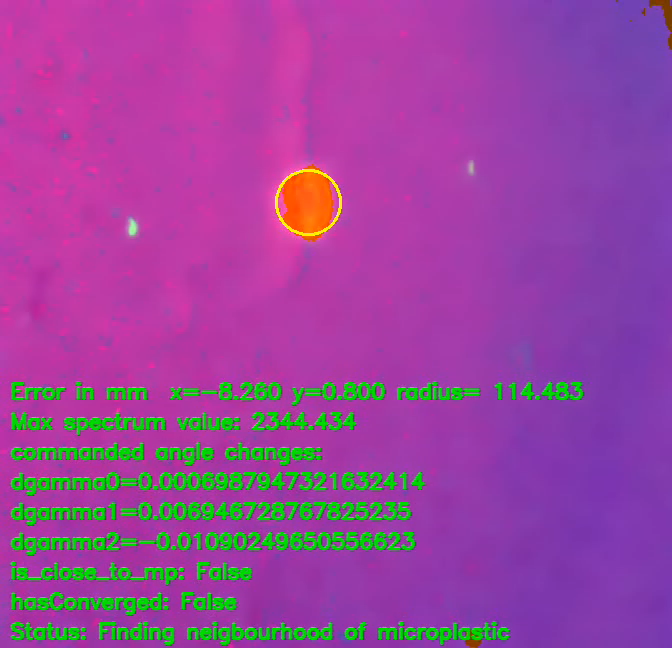}
        \caption*{Tracking mp}
    \end{minipage}
    \hfill
    \begin{minipage}{0.49\textwidth}
        \centering
        \includegraphics[width=\textwidth,trim=0 0 0 120,clip]{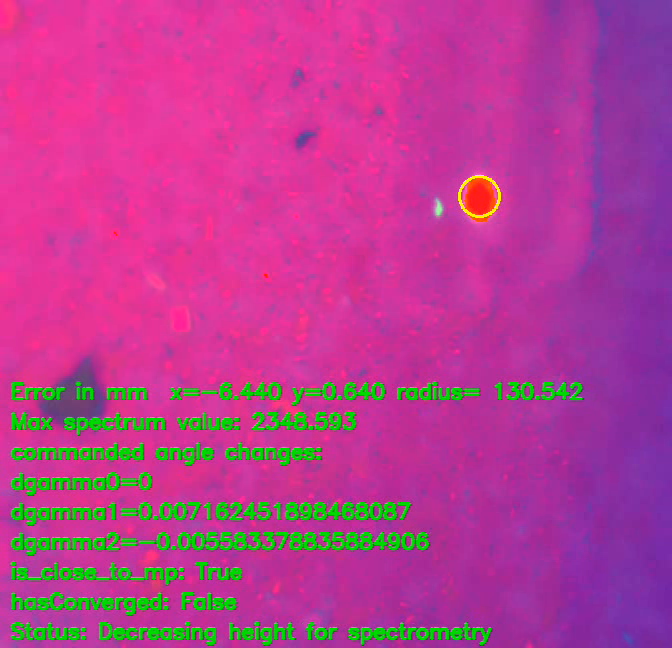}
        \caption*{Height adjustment}
    \end{minipage}
    \hfill
    \begin{minipage}{0.49\textwidth}
        \centering
        \includegraphics[width=\textwidth,trim=0 0 0 120,clip]{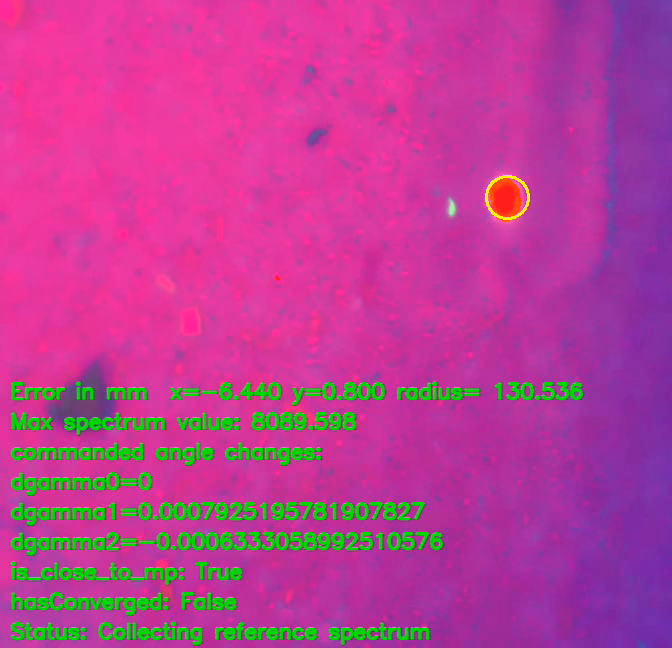}
        \caption*{Reference spectrum}
    \end{minipage}
    \hfill
    \begin{minipage}{0.49\textwidth}
        \centering
        \includegraphics[width=\textwidth,trim=0 0 0 120,clip,page=1]{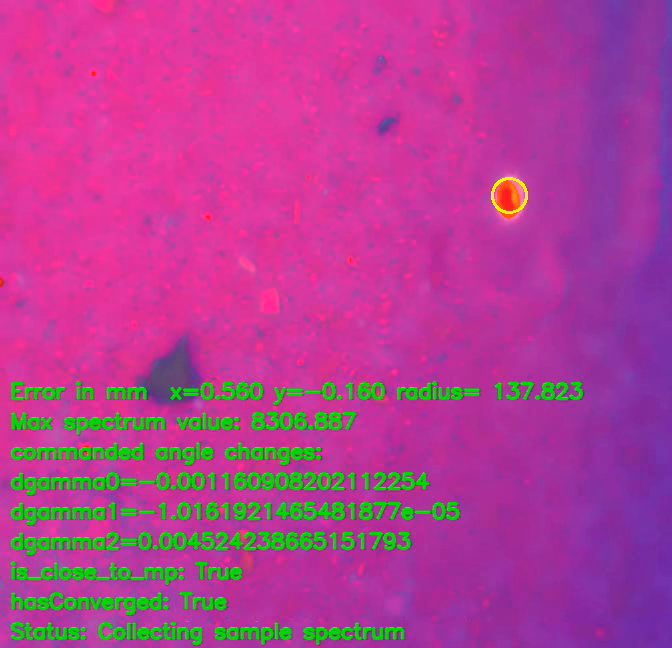}
        \caption*{Sample spectrum}
    \end{minipage}
    \caption{Sequential demonstration of the process.}
    \label{fig:demo-sequence2}
\end{figure*}

Tests show the sensitivity of NIR system to positional errors in Fig. \ref{fig:quantitative_spec}. We note that in order to get receive a valid absorbance of microplastic particle, less than 1 mm error in $z$ direction from the ``nominal'' pose is acceptable. Note also that depending on the size of microplastic particle, 1 mm error in $xy$ directions could result an invalid NIR spectra for very small plastic particles. 

To ensure adequate positional accuracy, we correct for kinematic errors in an endpoint closed-loop (EOL) fashion by observing \textit{both} end-effector and microplastic particle. However, as explained in section \ref{sec:related_work}, when the NIR sensors come close to the microplastic particle, the latter may no more be visible. This occlusion is handled with feedback of NIR measurements. As our signal processing and classifier (details in section \ref{sec:signal_processing}) is trained to detect both plastic and non-plastic materials, the NIR signal allows us to adjust the position of end effector in a ``teach by showing'' fashion, until the sensors sufficiently cover the microplastic particle surface \textit{and} an optimal SNR is observed. In Fig. \ref{fig:terminal_step}, we show the errors before and after the terminal control. For all of the trails, the terminal error remains close to or less than 1 mm. However, as our algorithm is designed to run in ``show by teaching'' fashion, a larger error can be iteratively corrected until a valid measurement is received.

A complete tracking process is demonstrated in Fig. \ref{fig:demo-sequence2} using snapshots from the eye-in-hand camera on mobile robot during a real-life experiment on beach. The NIR illumination diameter decreases as the end effector gets closer to the ground, before an optimal SNR is reached. Subsequently, reference NIR spectra is collected corresponding to the particular sand surface near (but not under) microplastic particle. Then terminal control is triggered to steer the NIR illumination on top of microplastic and converge even when the microplastic detection is lost due to occlusion. 



\subsection{Feature detection}
\label{sec:visual_tracking}

\begin{figure*}[ht]
    \centering
    
    \subfloat[]{\includegraphics[trim={\leftTrim} {\bottomTrim} {\rightTrim} {\topTrim}, clip, width=0.22\textwidth]{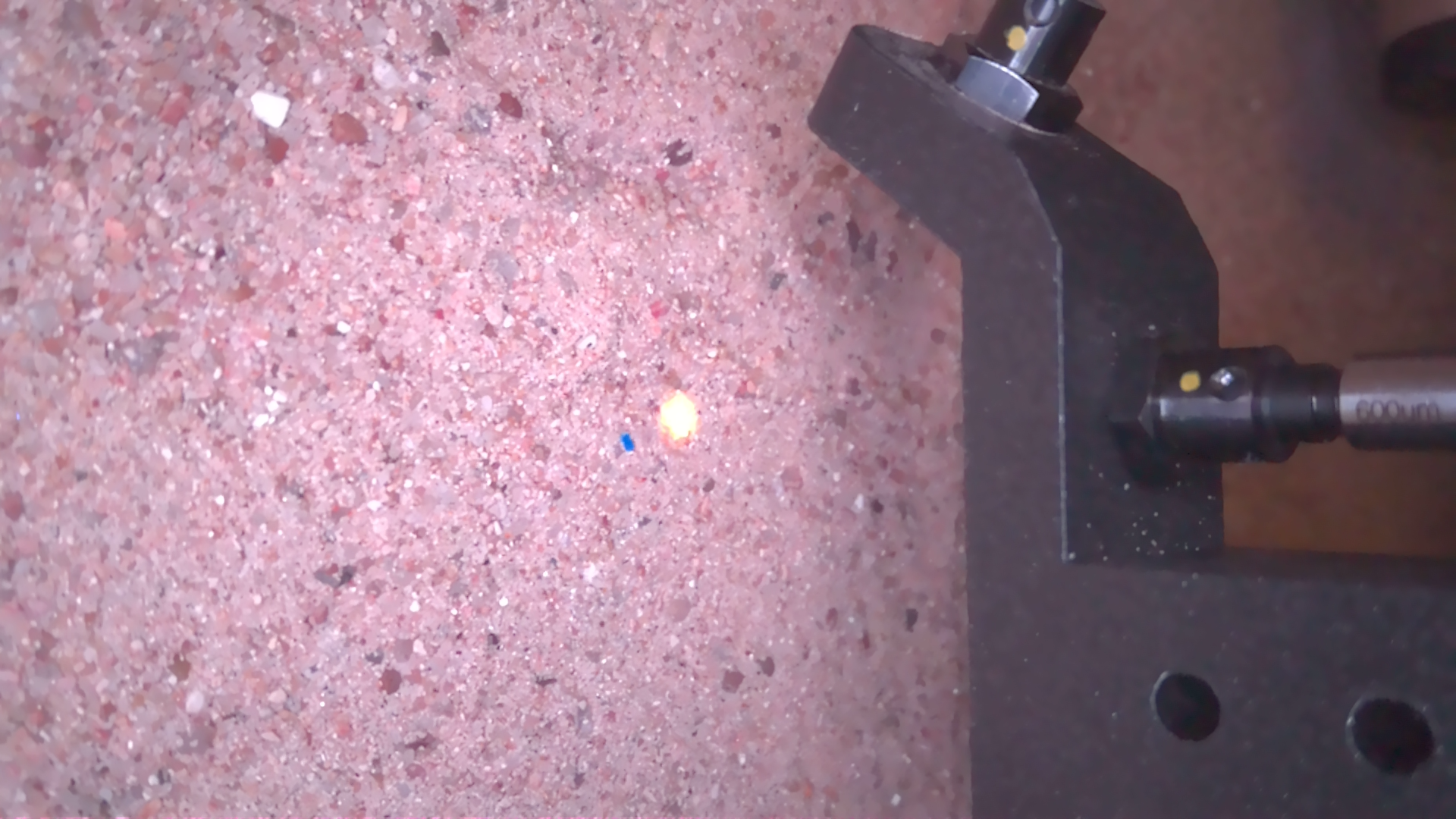}}%
    \hfill
    \subfloat[]{\includegraphics[trim={\leftTrim} {\bottomTrim} {\rightTrim} {\topTrim}, clip, width=0.22\textwidth]{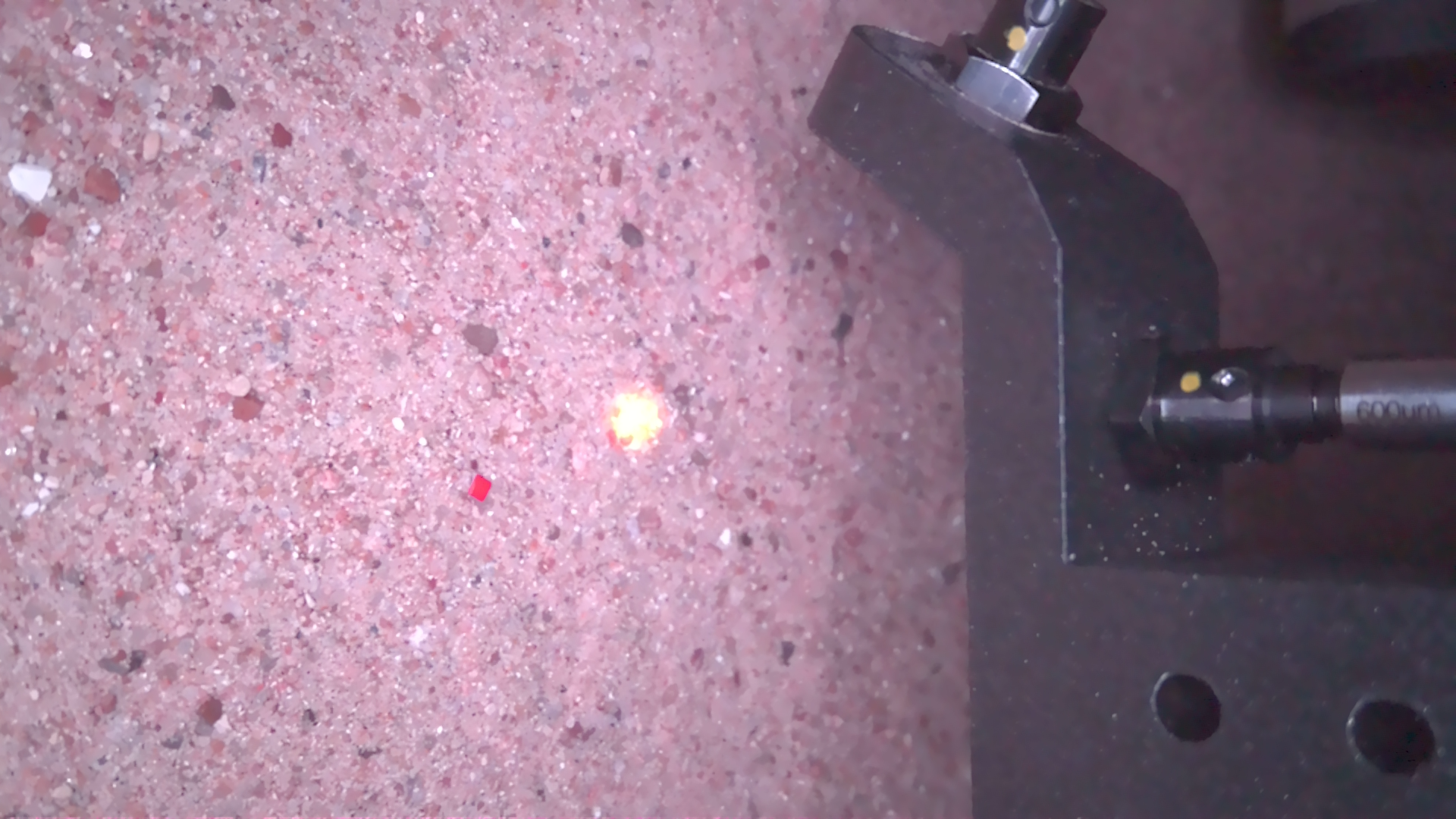}}%
    \hfill
    \subfloat[]{\includegraphics[trim={\leftTrim} {\bottomTrim} {\rightTrim} {\topTrim}, clip, width=0.22\textwidth]{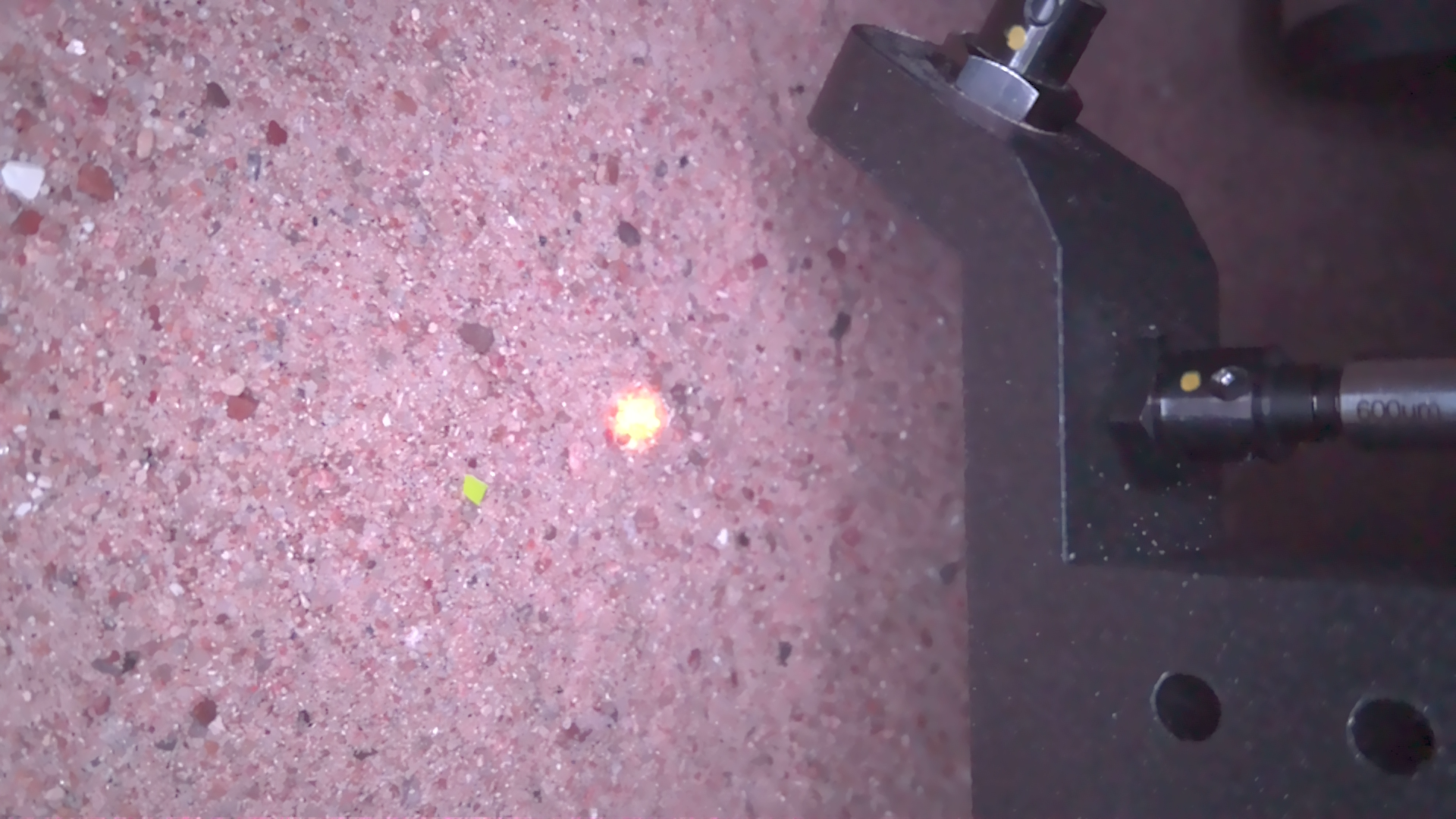}}%
    \hfill
    \subfloat[]{\includegraphics[trim={\leftTrim} {\bottomTrim} {\rightTrim} {\topTrim}, clip, width=0.22\textwidth]{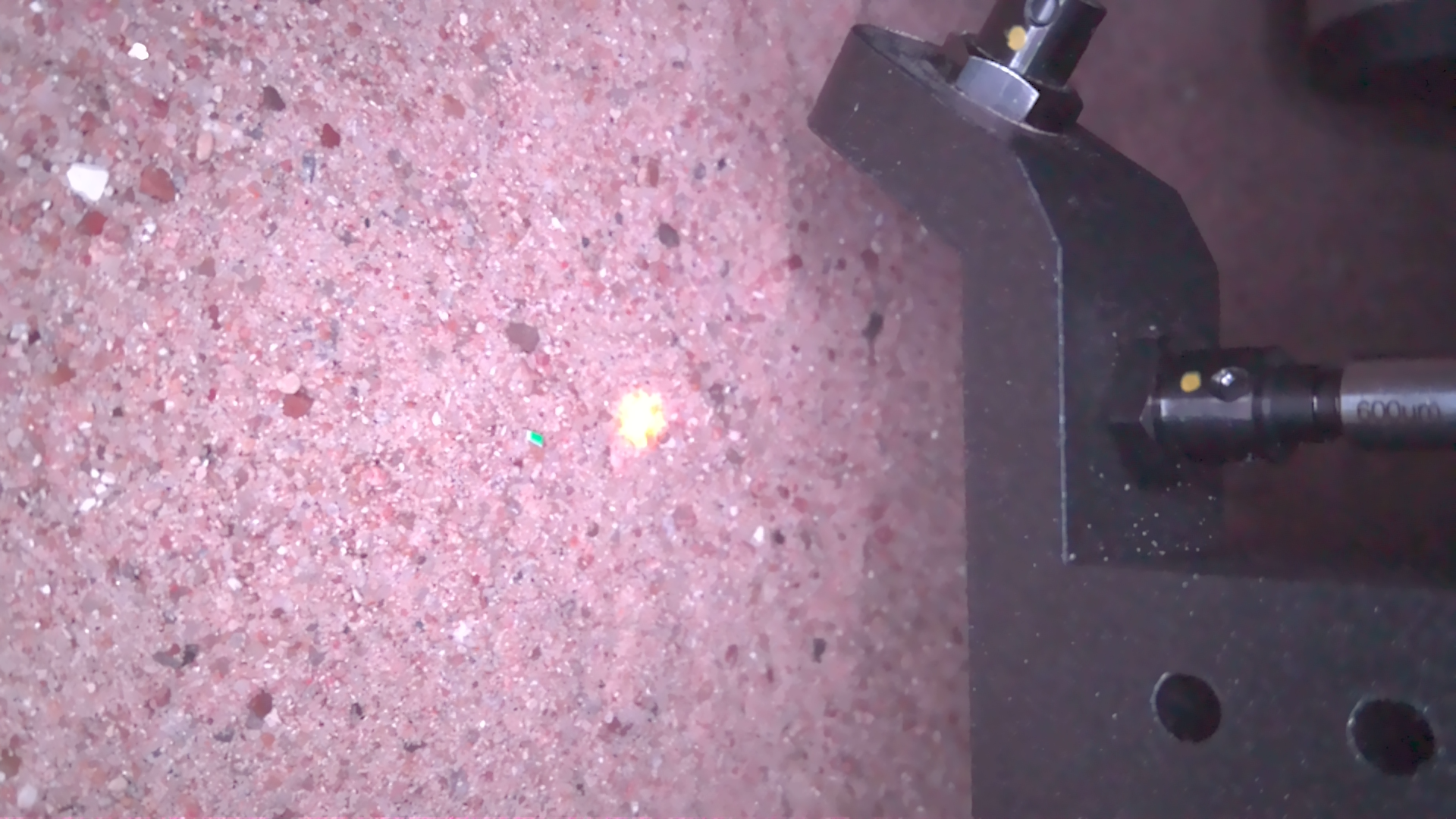}}%
    
    \vspace{0.5cm}
    
    \subfloat[]{\includegraphics[width=0.22\textwidth]{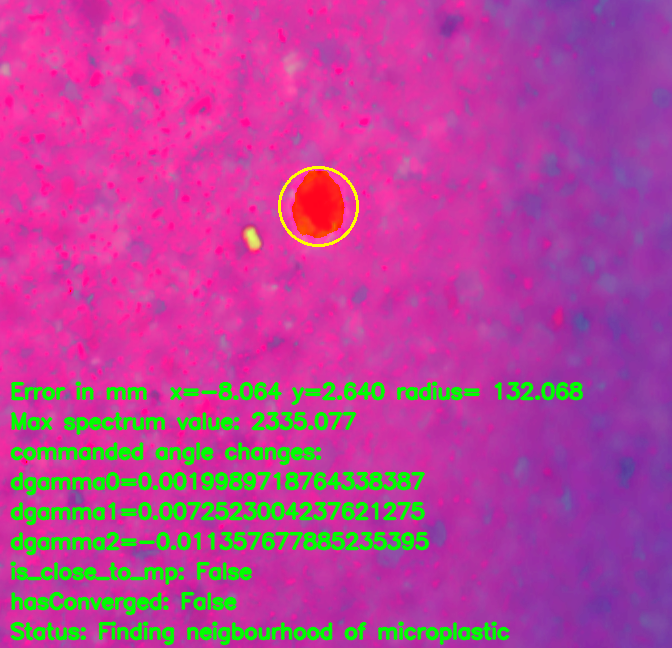}}%
    \hfill
    \subfloat[]{\includegraphics[width=0.22\textwidth]{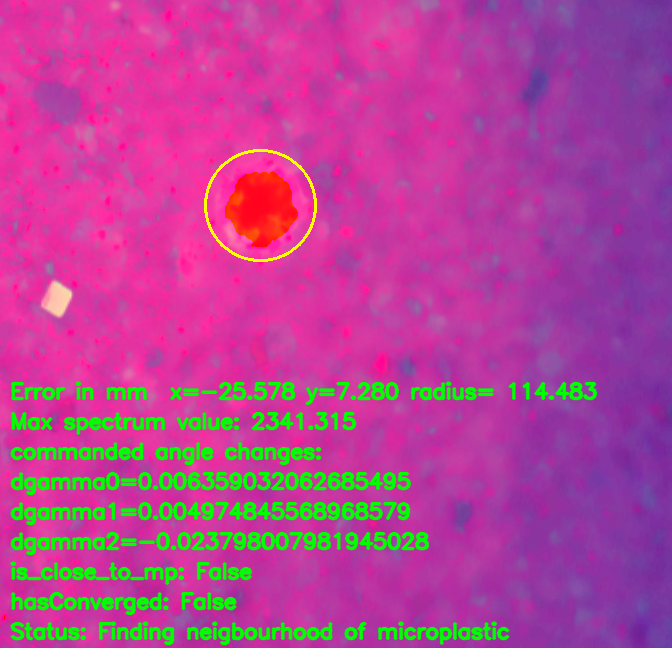}}%
    \hfill
    \subfloat[]{\includegraphics[width=0.22\textwidth]{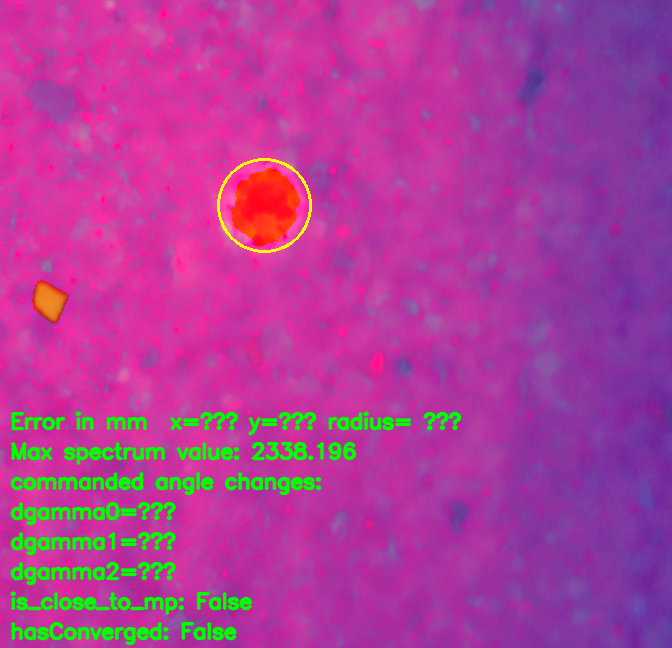}}%
    \hfill
    \subfloat[]{\includegraphics[width=0.22\textwidth]{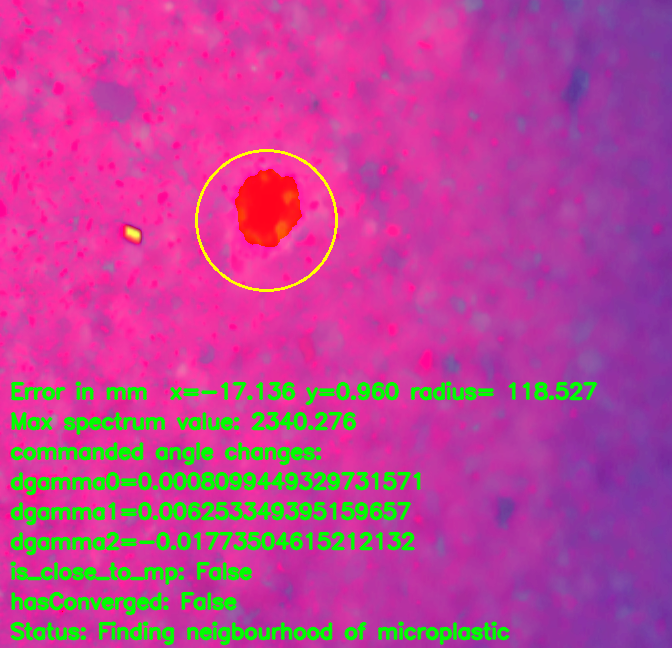}}%
    
     \caption{Saturation thresholding can detect microplastics of blue, red and green based colors. However, it fails to detect a microplastic of dim yellow color that is about the same color as sand.}
    \label{fig:thresholding}
\end{figure*}

In this section, we explain our approach to segment microplastics with background comprising of sand grains and organic matter like leaves, clams, plant debris, etc. As mentioned in section \ref{sec:related_work}, conventional machine learning models may be unreliable when class asymmetry is extremely stark i.e. false positives can dominate performance metrics when one class is overwhelmingly more common than others. It can be generally assumed that a typical beach surface may contain no more than a few hundred microplastic particles per $m^2$. The camera used in this work has FoV of 80\textdegree and its pose is maintained at about 10 cm away from the ground during scanning. Therefore, a microplastic particle of size $1\times1$ mm would occupy less than 0.01\% of all pixels. While large foundation models show promise in detecting objects, they require appropriate prompts that can only be identified in an ad-hoc manner. 

As our problem demands real-time decision-making in an unstructured environment, we rely on color, shape and size based thresholding and hierarchical contour analysis methods to segment candidate microplastics particles. Similarly, we segment NIR illumination spot on the ground using saturation based thresholding. Our tests show that we can successfully segment microplastics of varying colors including but not limited to blue, red and green. However, as shown in Fig \ref{fig:thresholding}, if the particle color is close to the sand color in terms of hue and saturation, our detection fails. We defer to future work the development of enhanced segmentation techniques capable of detecting microplastics across a broader range of colors, including transparent microplastics. 
\subsection{Signal processing and classification}
\label{sec:signal_processing}

We collect NIR spectra of plastic and non-plastic materials including PP, PET, PVC, PS, nylon, polylactic acid (PLA), high-density polyethylene (HDPE), low-density polyethylene (LDPE), rubber, cardboard, wood, bark and dry grass. We perturb the spectral data with Gaussian noise to capture different SNR levels. Such augmentation also increases the dataset size, as well as classification robustness, which is crucial in field operation where measurements are noisy and collimating lenses may not be perfectly aligned. 

During operation, we focus the NIR illumination onto microplastic sample using a ``emitter'' collimating lens. An identical ``receiver'' lens is oriented at 45\textdegree\ angle relative to emitter lens. Both lenses are attached to respective fiber optic cables for illuminating and collecting the remitted light. This geometry allows for sufficient collection of scattered light from plastic particle even when the plastic samples are rough, angled, rounded, or poorly positioned. For adaptive background subtraction, dark $R_\text{dark}$ and reference $R_\text{reference}$ spectra are taken before samples are measured in regular operation. We calculate the absorbance as

\begin{equation}
    A = -\log_{10} \frac{R_{\text{sample}} - R_{\text{dark}}}{R_{\text{reference}} - R_{\text{dark}}}
\end{equation}

\begin{figure*}
    \centering
    \begin{subfigure}[b]{0.48\textwidth}
        \centering
        \includegraphics[width=\textwidth]{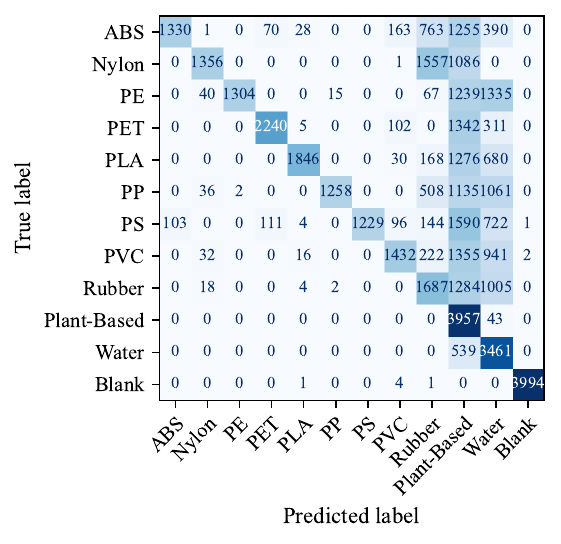}
    \end{subfigure}
    \hfill
    \begin{subfigure}[b]{0.48\textwidth}
        \centering
        \includegraphics[width=\textwidth]{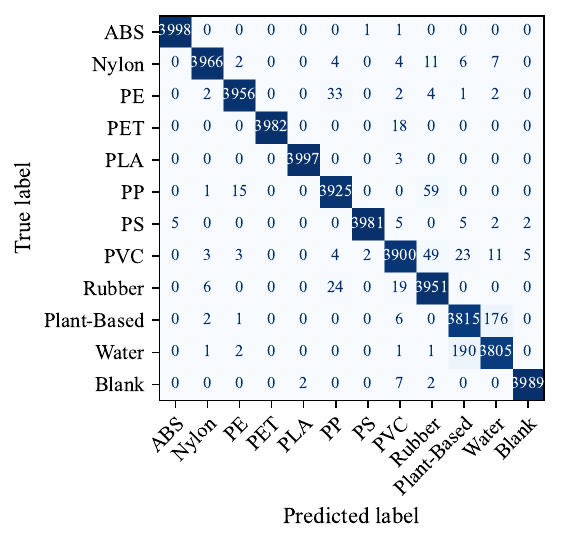}
    \end{subfigure}
    \caption{Confusion matrix for microplastics classification. Left: \texttt{SVM$^3$} model. Right: \texttt{SVM$^3$+I} model.}
    \label{fig:confusionmatrix}
\end{figure*}

Depending on the material, absorbance spectra is characterized by a series of overtone bands corresponding to the molecular vibrations of its constituent compounds. These overtone features occur at specific wavelengths and enable discrimination between non-plastic materials and various types of plastic polymers. Therefore, we train a classifier using absorbance spectra. In practice, the highest classification accuracy on both training and test datasets is achieved using a cubic support vector machine (\texttt{SVM}$^3$) model. The corresponding confusion matrix is shown in Fig. \ref{fig:confusionmatrix}. Notably, predictions on the test set indicate that training a different model (i.e. \texttt{SVM}$^3$+\texttt{I}) on augmented data that includes the so-called ``interferants'' \cite{shirley2025microplastics}, such as water and plant matter, in the background performs significantly better both in terms of reducing false positives and false negatives. We classify microplastics with the \texttt{SVM}$^3$+\texttt{I}) model during field experiments.  


\section{Discussion}

Our previous experiments demonstrate that this robotic platform is capable of detecting and chemically analyzing microplastics of size $\geq$ 1 mm with a high success rate. Our findings suggest that the proposed solution is fit to perform an extensive microplastic mapping on beaches. Currently, the robotic platform is programmed to autonomously navigate along the transect of hide tide line that is determine offline. In future, we plan to extend our exploration range with multi-robot system by devising informative path planning strategies \cite{du2021multi}. Furthermore, we defer to future work the task of assimilating spatiotemporal maps of microplastics distribution by chemical composition, and analysis of these maps to understand beach dynamics (source, accumulation zone identification as a function of grain size, etc.). 

Current microplastic segmentation relies on thresholding and contour processing, which has some limitations. For instance, if the microplastic is partially covered in sand, or the color is close to the sand color under LED illumination, the particle may go undetected. In future, we plan to improve upon segmentation by utilizing foundational models that can be run in real-time on AI-powered embedded computers. 

The off-the-shelf rover employed in this work has the maximum payload limit of 5 kg. This significantly limits options for robotic arms and end effectors. In future, we plan to develop a slightly larger rover, one that remains suitable for use in public beaches, yet can support a payload of 10-15 kg. This will allow manipulator control with improved tracking and convergence time. 

The microplastic type classifier proposed in this work is trained on data acquired from microplastics that are at least 1 mm in size. We introduced noise in the data to improve generalization and resilience to noisy measurements. However, false positives such as in neoprene rubber suggest the need for data collection at a wider spectral range. In future, we plan to improve classifier accuracy on smaller sized particles of 0.2 mm. 

\section{Conclusion}
In this paper, we proposed a mobile manipulator based automated approach to detect and chemically analyze microplastics on beach surfaces. We developed a state machine based control of the robotic platform to reach a set waypoint, scan the region with a camera mounted on end effector of robotic arm, and segment microplastics against backgrounds comprising sand particles and organic matter (leaves, clams, etc.). A new NIR and visual feedback based manipulator controller is proposed and shown to achieve $\leq$ 1 mm positional errors despite the occlusion that occurs when NIR illumination gets close to the microplastic particle. We measured a high number of spectra corresponding to diverse plastic and non-plastic materials, and trained a robust classifier that achieves up to 86.4\% accuracy during field operations. This work attempts to fill the gap in current methods within environmental exploration and autonomous science, greatly improving the speed at which beach environments can be mapperd, and minimizing human labor. We recommend further development of robotics-based solutions for real-time chemical analysis of microplastics, especially in more complex environments such as underwater marine ecosystems. 




\bibliographystyle{IEEEtran}
\bibliography{refs}

\begin{thebibliography}{10}
\providecommand{\url}[1]{#1}
\csname url@samestyle\endcsname
\providecommand{\newblock}{\relax}
\providecommand{\bibinfo}[2]{#2}
\providecommand{\BIBentrySTDinterwordspacing}{\spaceskip=0pt\relax}
\providecommand{\BIBentryALTinterwordstretchfactor}{4}
\providecommand{\BIBentryALTinterwordspacing}{\spaceskip=\fontdimen2\font plus
\BIBentryALTinterwordstretchfactor\fontdimen3\font minus
  \fontdimen4\font\relax}
\providecommand{\BIBforeignlanguage}[2]{{%
\expandafter\ifx\csname l@#1\endcsname\relax
\typeout{** WARNING: IEEEtran.bst: No hyphenation pattern has been}%
\typeout{** loaded for the language `#1'. Using the pattern for}%
\typeout{** the default language instead.}%
\else
\language=\csname l@#1\endcsname
\fi
#2}}
\providecommand{\BIBdecl}{\relax}
\BIBdecl

\bibitem{alvarez2020method}
J.~C. Alvarez-Zeferino, A.~A. Cruz-Salas, and A.~V{\'a}zquez-Morillas~et al.,
  ``Method for quantifying and characterization of microplastics in sand
  beaches,'' \emph{Revista internacional de contaminaci{\'o}n ambiental},
  vol.~36, no.~1, pp. 151--164, 2020.

\bibitem{graca2017sources}
B.~Graca, K.~Szewc, and D.~Zakrzewska~et al., ``Sources and fate of
  microplastics in marine and beach sediments of the southern baltic sea—a
  preliminary study,'' \emph{Environmental Science and Pollution Research},
  vol.~24, pp. 7650--7661, 2017.

\bibitem{dunbabin2012robots}
M.~Dunbabin and L.~Marques, ``Robots for environmental monitoring: Significant
  advancements and applications,'' \emph{IEEE Robotics \& Automation Magazine},
  vol.~19, no.~1, pp. 24--39, 2012.

\bibitem{angelopoulos2024transforming}
A.~Angelopoulos, J.~F. Cahoon, and R.~Alterovitz, ``Transforming science labs
  into automated factories of discovery,'' \emph{Science Robotics}, vol.~9,
  no.~95, p. eadm6991, 2024.

\bibitem{angelopoulos2023high}
A.~Angelopoulos, M.~Verber, and C.~McKinney~et al., ``High-accuracy injection
  using a mobile manipulation robot for chemistry lab automation,'' in
  \emph{2023 IEEE/RSJ International Conference on Intelligent Robots and
  Systems (IROS)}.\hskip 1em plus 0.5em minus 0.4em\relax IEEE, 2023, pp.
  10\,102--10\,109.

\bibitem{sauvee2008ultrasound}
M.~Sauv{\'e}e, P.~Poignet, and E.~Dombre, ``Ultrasound image-based visual
  servoing of a surgical instrument through nonlinear model predictive
  control,'' \emph{The International Journal of Robotics Research}, vol.~27,
  no.~1, pp. 25--40, 2008.

\bibitem{ma2024guiding}
X.~Ma, M.~Zeng, and J.~C. Hill~et al., ``Guiding the last centimeter: Novel
  anatomy-aware probe servoing for standardized imaging plane navigation in
  robotic lung ultrasound,'' \emph{IEEE Transactions on Automation Science and
  Engineering}, 2024.

\bibitem{mebarki20102}
R.~Mebarki, A.~Krupa, and F.~Chaumette, ``2-d ultrasound probe complete
  guidance by visual servoing using image moments,'' \emph{IEEE Transactions on
  Robotics}, vol.~26, no.~2, pp. 296--306, 2010.

\bibitem{nadeau2013intensity}
C.~Nadeau and A.~Krupa, ``Intensity-based ultrasound visual servoing: Modeling
  and validation with 2-d and 3-d probes,'' \emph{IEEE Transactions on
  Robotics}, vol.~29, no.~4, pp. 1003--1015, 2013.

\bibitem{nadeau2016moments}
C.~Nadeau, A.~Krupa, J.~Petr, and C.~Barillot, ``Moments-based ultrasound
  visual servoing: From a mono-to multiplane approach,'' \emph{IEEE
  Transactions on Robotics}, vol.~32, no.~6, pp. 1558--1564, 2016.

\bibitem{pane2022ultrasound}
S.~Pane, G.~Faoro, and E.~Sinibaldi~et al., ``Ultrasound acoustic phase
  analysis enables robotic visual-servoing of magnetic microrobots,''
  \emph{IEEE Transactions on Robotics}, vol.~38, no.~3, pp. 1571--1582, 2022.

\bibitem{baumgartner2006mobile}
E.~T. Baumgartner, R.~G. Bonitz, and J.~P. Melko~et al., ``Mobile manipulation
  for the mars exploration rover-a dexterous and robust instrument positioning
  system,'' \emph{IEEE robotics \& automation magazine}, vol.~13, no.~2, pp.
  27--36, 2006.

\bibitem{bowkett2024challenges}
J.~Bowkett, M.~Dolci, and J.~Aldrich~et al., ``Challenges in closed-loop
  compliant motion control for planetary robotics,'' in \emph{2024 IEEE
  Aerospace Conference}.\hskip 1em plus 0.5em minus 0.4em\relax IEEE, 2024, pp.
  1--11.

\bibitem{backes2005automated}
P.~Backes, A.~Diaz-Calderon, and M.~Robinson~et al., ``Automated rover
  positioning and instrument placement,'' in \emph{2005 IEEE Aerospace
  Conference}.\hskip 1em plus 0.5em minus 0.4em\relax IEEE, 2005, pp. 60--71.

\bibitem{fleder2011autonomous}
M.~Fleder, I.~A. Nesnas, and M.~Pivtoraiko~et a., ``Autonomous rover traverse
  and precise arm placement on remotely designated targets,'' in \emph{2011
  IEEE International Conference on Robotics and Automation}.\hskip 1em plus
  0.5em minus 0.4em\relax IEEE, 2011, pp. 2190--2197.

\bibitem{hanson2023slurp}
N.~Hanson, W.~Lewis, and K.~Puthuveetil~et al., ``Slurp! spectroscopy of
  liquids using robot pre-touch sensing,'' in \emph{2023 IEEE International
  Conference on Robotics and Automation (ICRA)}.\hskip 1em plus 0.5em minus
  0.4em\relax IEEE, 2023, pp. 3786--3792.

\bibitem{hanson2024prospect}
N.~Hanson, G.~Lvov, and V.~Rautela~et al., ``Prospect: Precision robot
  spectroscopy exploration and characterization tool,'' in \emph{2024 IEEE/RSJ
  International Conference on Intelligent Robots and Systems (IROS)}.\hskip 1em
  plus 0.5em minus 0.4em\relax IEEE, 2024, pp. 5244--5251.

\bibitem{huang2023proceeding}
H.~Huang, H.~Cai, and J.~U. Qureshi~et al., ``Proceeding the categorization of
  microplastics through deep learning-based image segmentation,'' \emph{Science
  of the Total Environment}, vol. 896, p. 165308, 2023.

\bibitem{park2022mp}
H.-m. Park, S.~Park, and M.~K. de~Guzman~et al., ``Mp-net: Deep learning-based
  segmentation for fluorescence microscopy images of microplastics isolated
  from clams,'' \emph{Plos one}, vol.~17, no.~6, p. e0269449, 2022.

\bibitem{xu2024efficient}
J.~Xu and Z.~Wang, ``Efficient and accurate microplastics identification and
  segmentation in urban waters using convolutional neural networks,''
  \emph{Science of The Total Environment}, vol. 911, p. 168696, 2024.

\bibitem{phan2023exploiting}
S.~Phan, D.~Torrejon, and J.~Furseth~et al., ``Exploiting weak supervision to
  facilitate segmentation, classification, and analysis of microplastics (< 100
  $\mu$m) using raman microspectroscopy images,'' \emph{Science of the Total
  Environment}, vol. 886, p. 163786, 2023.

\bibitem{shirley2025microplastics}
J.~C. Shirley, K.~A. Rex, and H.~Iqbal~et al., ``Microplastics in the rough:
  Using data augmentation to identify plastics contaminated by water and plant
  matter,'' \emph{RSC Sustainability}, 2025.

\bibitem{dong2016incremental}
G.~Dong and Z.~H. Zhu, ``Incremental visual servo control of robotic
  manipulator for autonomous capture of non-cooperative target,''
  \emph{Advanced Robotics}, vol.~30, no.~22, pp. 1458--1465, 2016.

\bibitem{kelly2000stable}
R.~Kelly, R.~Carelli, and O.~Nasisi~et al., ``Stable visual servoing of
  camera-in-hand robotic systems,'' \emph{IEEE/ASME transactions on
  mechatronics}, vol.~5, no.~1, pp. 39--48, 2000.

\bibitem{wijesoma1993eye}
S.~Wijesoma, D.~Wolfe, and R.~Richards, ``Eye-to-hand coordination for
  vision-guided robot control applications,'' \emph{The International Journal
  of Robotics Research}, vol.~12, no.~1, pp. 65--78, 1993.

\bibitem{yoshida2022automated}
T.~Yoshida, Y.~Onishi, and T.~Kawahara~et al., ``Automated harvesting by a
  dual-arm fruit harvesting robot,'' \emph{ROBOMECH journal}, vol.~9, no.~1,
  p.~19, 2022.

\bibitem{bajracharya2006vision}
M.~Bajracharya, M.~DiCicco, and P.~Backes, ``Vision-based end-effector position
  error compensation,'' in \emph{2006 IEEE Aerospace Conference}.\hskip 1em
  plus 0.5em minus 0.4em\relax IEEE, 2006, pp. 7--pp.

\bibitem{sun2024efficient}
T.~Sun, W.~Zhang, and X.~Gao~et al., ``Efficient occlusion avoidance based on
  active deep sensing for harvesting robots,'' \emph{Computers and Electronics
  in Agriculture}, vol. 225, p. 109360, 2024.

\bibitem{border2024surface}
R.~Border and J.~D. Gammell, ``The surface edge explorer (see): A
  measurement-direct approach to next best view planning,'' \emph{The
  International Journal of Robotics Research}, vol.~43, no.~10, pp. 1506--1532,
  2024.

\bibitem{radmard2018resolving}
S.~Radmard, D.~Meger, J.~J. Little, and E.~A. Croft, ``Resolving occlusion in
  active visual target search of high-dimensional robotic systems,'' \emph{IEEE
  Transactions on Robotics}, vol.~34, no.~3, pp. 616--629, 2018.

\bibitem{gupta2025training}
A.~Gupta, R.~Sathua, and S.~Gupta, ``A training-free framework for precise
  mobile manipulation of small everyday objects,'' \emph{arXiv preprint
  arXiv:2502.13964}, 2025.

\bibitem{xin2021visual}
J.~Xin, H.~Cheng, and B.~Ran, ``Visual servoing of robot manipulator with weak
  field-of-view constraints,'' \emph{International Journal of Advanced Robotic
  Systems}, vol.~18, no.~1, p. 1729881421990320, 2021.

\bibitem{hutchinson1996tutorial}
S.~Hutchinson, G.~D. Hager, and P.~I. Corke, ``A tutorial on visual servo
  control,'' \emph{IEEE transactions on robotics and automation}, vol.~12,
  no.~5, pp. 651--670, 1996.

\bibitem{du2021multi}
B.~Du, K.~Qian, and H.~Iqbal~et al., ``Multi-robot dynamical source seeking in
  unknown environments,'' in \emph{2021 IEEE International Conference on
  Robotics and Automation (ICRA)}.\hskip 1em plus 0.5em minus 0.4em\relax IEEE,
  2021, pp. 9036--9042.

\end{thebibliography}

\end{document}